\documentclass[lettersize,journal]{IEEEtran}
\usepackage[dvipsnames, table]{xcolor}
\usepackage{amsmath,amsfonts, nccmath}
\usepackage{algorithmic}
\usepackage{array}
\usepackage[caption=false,font=footnotesize]{subfig}
\usepackage{textcomp}
\usepackage{stfloats}
\usepackage{url}
\usepackage{verbatim}
\usepackage{graphicx}
\usepackage{cite}
\usepackage{booktabs}
\usepackage[vlined,linesnumbered,ruled,norelsize]{algorithm2e}
\SetKwInOut{Input}{input}\SetKwInOut{Output}{output}
\usepackage[inline]{enumitem}
\usepackage{tabularx, booktabs,ragged2e, multirow}
\usepackage{tikz}
\usetikzlibrary {positioning, arrows, calc, arrows.meta, patterns, shapes.geometric, bending, shadings, shapes.arrows, decorations.text, shapes.callouts}
\usepackage{amsthm}

\theoremstyle{plain}

\theoremstyle{remark}
\newtheorem{remark}{Remark}

\begin{document}

\title{To Lead or to Follow? Adaptive Robot Task Planning in Human-Robot Collaboration}


\author{Ali Noormohammadi-Asl,~\IEEEmembership{Member,~IEEE}, Stephen L. Smith,~\IEEEmembership{Senior Member,~IEEE}, \\ and Kerstin Dautenhahn,~\IEEEmembership{Fellow,~IEEE}
\thanks{This research was undertaken, in part, thanks to funding from the Natural Sciences and Engineering Research Council of Canada (NSERC) and the Canada 150 Research Chairs Program.

Ali~Noormohammadi~Asl, Stephen~L.~Smith, and Kerstin~Dautenhahn are with the Department of Electrical and Computer Engineering, Faculty of Engineering, University of Waterloo, Waterloo, ON N2L
3G1, Canada (e-mail: ali.asl@uwaterloo.ca, stephen.smith@uwaterloo.ca, kerstin.dautenhahn@uwaterloo.ca).}
}



\maketitle

\begin{abstract}
Adaptive task planning is fundamental to ensuring effective and seamless human-robot collaboration. This paper introduces a robot task planning framework that takes into account both human leading/following preferences and performance, specifically focusing on task allocation and scheduling in collaborative settings. We present a proactive task allocation approach with three primary objectives: enhancing team performance, incorporating human preferences, and upholding a positive human perception of the robot and the collaborative experience. Through a user study, involving an autonomous mobile manipulator robot working alongside participants in a collaborative scenario, we confirm that the task planning framework successfully attains all three intended goals, thereby contributing to the advancement of adaptive task planning in human-robot collaboration. This paper mainly focuses on the first two objectives, and we discuss the third objective, participants' perception of the robot, tasks, and collaboration in a companion paper.

\end{abstract}


\section{Introduction}
\IEEEPARstart{C}{obots}, short for collaborative robots, have signified a transformative leap from traditional industrial robots, working isolated from humans, to robots that can share their workspace with their human coworkers, laying the ground to exploit the synergy of human-robot collaboration (HRC). Although cobots are currently slower and less powerful than traditional industrial robots, mainly due to their proximity to humans and safety concerns, they are easy to install and relocate and are productive and cost-effective automation solutions for diverse work environments, even small enterprises \cite{vicentini2021collaborative}. Leveraging these capabilities and establishing a seamless collaboration, however, can be achieved only through human-aware programming of cobots \cite{el2019cobot}, empowering them to learn, adapt, and work robustly. 

One of the main challenges in programming cobots is enabling them to adapt to their human teammate, especially their preferences. This topic has been extensively researched in the context of human-robot interaction (HRI) and collaboration, primarily centering on considering human preferences and enhancing human satisfaction and perception of the robot\cite{mitsunaga2008adapting, nemlekar2023transfer, wang2018facilitating, grigore2018preference}. Clearly, a high level of human perception of the robot facilitates effective long-term collaboration between humans and cobots, but a question remains: ``Do only the human teammate's satisfaction and perception matter?". Cobots are typically less expensive compared to traditional industrial robots, yet they still need to demonstrate sufficient productivity to convince business owners to invest in them \cite{el2019cobot}. Hence, team performance, in addition to human perception of the robot and collaboration, are two important factors to consider in cobot programming. 
\begin{figure}[!t]
\centering
\scalebox{0.45}{
\input{pic/framework_new}}
\caption{{Overview of the robot planning framework. The robot adapts its task planning based on task goals (e.g., completion time), and its estimates of the human's leading/following preferences and performance. The framework continuously observes human actions and updates its estimates accordingly, allowing the robot to dynamically adjust its role and task allocation decisions.} }
\label{fig:fol_lead_framework}
\end{figure} 

Both the human and robot contribute to team performance with their complementary skill, but in the context of task planning and scheduling, the robot's high computational and planning abilities allow it to take a more significant and leading role, ensuring good team performance in the short term. However, due to human presence and dynamic environment uncertainties, it is sometimes more efficient for the human interaction partner to plan for the team \cite{el2019cobot}. However, the objectives of maximizing team performance and human perception may be conflicting. This leads to the following problem, which is the focus of this paper:
\textit{How do we enable a cobot to adapt to its human coworker's preferences to optimize human perception while keeping the team performance at an acceptable level?}

\IEEEpubidadjcol

We previously conducted an online user study involving a scenario with a single human and a robot, requiring the two parties to collaborate to accomplish a given task \cite{noormohammadi2021effect}. We considered three robot strategies: prioritizing the human (Strategy 1), prioritizing the robot (Strategy 2), and balancing both (Strategy 3). Based on the results, Strategies 1 and 3 enhanced human perception of the collaboration compared to Strategy 2, with no significant difference between Strategies 1 and 3. Those previous findings proposed that having a balanced robot plan in collaborative scenarios benefits the team without diminishing human perception of the collaboration. In this prior work, although the robot’s decisions were based on the state of the task, there was no adaptation to human preferences, and only fixed strategies were considered.

Resting on the insights gathered from this initial study, we have proposed a framework that equips the robot to estimate and adapt its planning to 
\begin{enumerate*}
    \item the human leading/following preferences, and 
    \item the human performance. 
\end{enumerate*}

In this framework, in contrast to most common approaches, which strictly focus on one end of the leading/following spectrum, the cobot gradually and continuously adapts to human performance and (leading/following) preferences in pursuit of long-term performance and can encompass the entire spectrum between acting solely as a leader or solely a follower. Fig.~\ref{fig:fol_lead_framework} encapsulates the main idea of this framework. 

The framework allows the robot to monitor and adapt to changes in human preferences and performance during the collaboration rather than relying on fixed, long-term preferences and performance. For example, consider a person with a following preference who becomes fatigued at some point and then prefers to assign more tasks to the robot, or a scenario in which a person with a high level of performance becomes confused and makes a few mistakes. This framework can be applied to different collaborative scenarios; however, our research focuses specifically on task selection/allocation and scheduling in human-robot collaboration.

In a basic task allocation problem, agents need to be assigned tasks with associated payoffs, aiming to optimize overall team payoff. This is a widely explored topic in multi-robot teams \cite{Khamis2015}, as well as HRC, involving a blend of human and robot agents\cite{schmidbauer2023empirical, pupa2022resilient, cheng2021human, lamon2019capability, darvish2020hierarchical}. Our work focuses on a single-human, single-robot teams, and what makes it different from conventional task allocation systems is: \emph{the human and cobot's agency in selecting their own tasks and assigning tasks to each other instead of being assigned merely by one of the agents or a third party (e.g., a manager/central controller).}

The provided agency allows the human to demonstrate and implement their leading/following preference in their collaboration with the cobot. Subsequently, throughout the collaboration, the cobot needs to estimate the latent human preference, monitor and estimate the human's performance, and adapt its planning online. The robot performs two-step task planning at each step: first, task allocation considering its belief about human preference and performance, and second, task scheduling. In this framework, the robot needs to re-assume the leading role when the human's performance is poor, even if the human agent prefers to lead the team.

In \cite{noormohammadi2022task}, we tested the framework in a simulation environment using a simplified model of human decision-making. In \cite{noormohammadi2023adapting}, we also implemented it on an actual mobile manipulator, the Fetch robot, and tested it by having the experimenter enact some possible different human collaboration styles. However, to test the system and planning framework's effectiveness and their influence on human perception of the robot, we conducted a user study, which is the primary focus of this paper. We design a collaborative scenario, inspired by the kitting task, with a set of precedence-constrained tasks that must be completed through collaboration between the human and the cobot. Both agents make their decisions asynchronously and can select tasks for themselves or assign them to each other.

\subsection{Contributions}
Restating the problem that we aim to tackle in this paper, adapting to human leading/following preferences while maintaining team performance at a high level, the main contributions of this paper are as follows:

\begin{enumerate}
    \item We propose a robot planning framework enabling the robot to consider both human leading/following preference and performance simultaneously. We apply this framework  specifically to the task allocation problem and present a two-step planning structure: 1- Task allocation by considering the robot's belief about the human agent's performance and leading/following preference, and 2- task scheduling.

    \item Our planning algorithm dynamically updates task states based on both agents' actions and actively identifies and addresses errors made by the human agent.

    \item We present the development and practical implementation of adaptive robot task planning in a collaborative scenario involving a robot performing autonomous pick-and-place. 

    \item Through a comprehensive user study with 48 participants, we demonstrate that the planning framework empowers the robot to proactively infer participants' performance and leading/following preferences in its task planning. The study also reveals the framework's ability to adapt to changes, such as participants preferring to follow the robot in challenging tasks or the robot reassuming the leading role when participants' performance decreases.
\end{enumerate}

Preliminary versions of parts of this work appeared in the conference papers~\cite{noormohammadi2022task, noormohammadi2023adapting}.  In~\cite{noormohammadi2022task} we introduced a planning framework based on human preference and performance, limited to a collaborative scenario in a simulation environment with a simplified human decision-making model. In contrast, this paper applies the framework to a more complex collaborative scenario involving an actual robot working alongside recruited participants.
In~\cite{noormohammadi2023adapting} we presented an initial version of implementing this collaborative scenario and assessed the planning framework's performance for four different scenarios conducted by the experimenter. However, the current paper presents the final version of the experimental setup with modifications to the planning framework to minimize frequent changes in robot planning. Importantly, it evaluates the efficiency of the proposed planning framework for an autonomous robot collaborating individually with each of the 48 participants (involving in total 144 tasks) and discusses specific cases to demonstrate the framework's adaptability to different participant preferences and performance levels and their variation.

We also note that due to the broad scope of the planning framework and user study, we have written this paper to focus primarily on aspects of robot planning, both theory and user study evaluation.  We have then written a companion paper~\cite{fetch_human}, to focus on participants' perceptions of the robot, tasks, and collaboration, providing insight into their actions within this collaborative context.

The remainder of this paper is structured as follows. Section II provides a review of relevant literature on task allocation and adaptation within the context of HRI and HRC. Section III presents the problem statement and introduces the proposed framework. Section IV delves into the study's design, and the implementation of the planning and estimation method, and outlines the study procedure. In Section V, we initially summarized the results pertaining to human perception of the robot, collaboration, and tasks. Subsequently, we analyze the results, focusing on the robot's planning and its effectiveness in adapting to participants. Finally, Section VI concludes the paper,

\section{Related Work}

In this section, we initially delve into related research concerning task allocation and the incorporation of human preferences into task allocation and planning. 

\subsection{Task Allocation}
Task allocation in HRC involves a suitable allocation of tasks to the human and robot agents and finding a proper chronological order for completing tasks based on problem-dependent decision factors and constraints. We can categorize task allocation approaches into two main groups: offline and online.

\subsubsection{Offline Task Allocation}
In offline task allocation, typically, the goal is to assign tasks based on prior knowledge of the suitability of agents for each task. As a measure of suitability, one can consider the proportionality of agents' abilities and constraints to the tasks' requirements and constraints. After determining suitability measures, the task allocation can be done by an expert \cite{schmidbauer2020adaptive, muller2016process}, via simulation studies\cite{tsarouchi2017human, michalos2018method}, or through mathematical modeling and optimization \cite{lamon2019capability, lee2022task}. In \cite{tsarouchi2017human}, after deciding possible suitable agents for each task, different cases are evaluated by simulation, and the one with the best utility value is selected. Similarly, in  \cite{michalos2018method}, multiple criteria are calculated through simulations, and a depth-first search algorithm is employed to find an optimal solution. In \cite{lamon2019capability}, the authors designed a human capability-based cost function to minimize human risk factors. Then, they applied their method in a user study,  using the A\textsuperscript{*} algorithm for role assignment. In \cite{lee2022task}, the problem of disassembly sequence planning is formulated as an optimization problem to minimize the disassembly time while considering resource and safety constraints.

\subsubsection{Online Task Allocation}
Offline task allocation approaches force the human-robot team to adhere to the (optimal) plan obtained offline. However, in many real-world scenarios, uncertainties that might arise due to individual preferences and behaviors, alterations in the workspace, and changes in task requirements question the applicability of these approaches. Online task allocation methods aim to cope with this limitation  by endowing the system with online replanning abilities. Typically, these methods fall into two groups. The first group of approaches, similar to the offline methods, finds an optimal task allocation, and the agents must follow the obtained plan. However, these methods are able to reactively replan online when the current task allocation is not valid anymore \cite{faroni2023optimal, cheng2021human, pupa2022resilient, alirezazadeh2022dynamic}.

The second group of methods mainly relies on the human agents' decisions, and the robot agents play more of a supportive role by adapting to the human agents. These methods do not create a fixed allocation requiring agents to adhere to it. In these methods, the human agents can make their own decisions and manage arising uncertainties, and the robot agents need to adapt their decisions proactively. In \cite{nemlekar2021two}, the robot infers the human preference by a two-stage clustering approach and provides the parts for the human in an assembly task. Other work proposed a real-time decision-making mechanism for a cobot based on the human's short-term and long-term behaviors \cite{gorur2023fabric}. In \cite{fiore2016planning}, the robot infers human preferences and chooses one of the three operation modalities: 1- the robot plans, 2- the human plans, and 3- the robot adapts.

\subsubsection{Human Preferences in Task Allocation}
As discussed in the preceding paragraph, it is crucial for the robot to proactively infer human preferences when taking on a supportive role. However, incorporating human preferences in situations where the robot or a central unit (e.g., a manager, central computer, etc) is actively involved in task allocation and planning is an aspect that is often underestimated. This aspect, central to our research, has been demonstrated as a critical element in enhancing humans' positive perception of the robot \cite{tausch2022best, gombolay2017computational, fiore2016planning, wilcox2013optimization}. In \cite{tausch2022best}, different scenarios were explored in which the manager, the robot, or the participants themselves assign tasks. In \cite{gombolay2017computational}, {human participants’ task type preferences are considered while balancing workloads, but during offline planning and often through direct inquiry.}
In \cite{fiore2016planning}, the system can adjust to user preferences, enabling the human partner to issue commands, or take a more passive role and seamlessly transition between different modes as needed. {The authors in \cite{wilcox2013optimization} propose a robust scheduling method that responds to temporal disturbances and adapts to changes in human task preferences during collaboration.} 

What is notably missing is an assessment of participants' performance to determine whether their preferences and performance align with the team's overall performance or potentially hinder it. {In \cite{liu2021coordinating}, the authors present a robust and adaptive task scheduling framework that models human task performance and learning curves under uncertainty. However, this approach does not consider human preferences. Additionally, much of the prior work assumes robot-led task scheduling without shared control, leaving the human agent uninvolved in task allocation.}

\subsection{Adaptation}
As noted in the task allocation literature, robot adaptation is an integral ability for a robot to collaborate smoothly and efficiently with humans. This adaptation can be achieved through manually teaching the robot by experts (i.e., how to collaborate effectively while adhering to human preferences) \cite{akgun2012keyframe, huang2019synthesizing}, equipping the robot with the ability to learn and adapt itself autonomously, or a combination of both.  Manual instruction by experts, while fast and effective, faces practical limitations, including time consumption, challenges in conveying nuanced instructions, and scalability issues \cite{unhelkar2020effective}. 

Learning-based adaptation approaches typically employ supervised or unsupervised learning-based algorithms. The latter involves identifying decision factors (features) influencing human behavior and collecting behavioral data, which is then enriched through expert annotation or participant surveys\cite{ayoub2023real, soh2020multi, huang2021meta}. The former, however, enables machines or robots to learn human preferences by observing their behavior and actions without the explicit need for annotation or data labeling\cite{luo2018unsupervised, nemlekar2021two,reddy2022first, kanazawa2019adaptive}. Some research also leverages both learning methods and experts' knowledge \cite{unhelkar2020effective, bajcsy2018learning}.

However, these studies often emphasize the robot's adaptation to the human agent. In contrast, mutual adaptation — where both the human and robot adjust based on each other's actions and feedback — can be essential in human-robot collaboration. The authors of \cite{nikolaidis2017human} explored mutual adaptation between humans and robots, where the robot initially guides the adaptable human but may unilaterally adapt if the human insists on their poor or inadequate performance or decisions.

\subsection{Being in or under control}

    The concept of control—whether being in control or under control—has long been discussed across various fields, including social sciences, human-robot (or computer) interaction, and industrial automation. In the context of human-robot interaction and automation, human perception of control, or the level of autonomy, can be influenced by factors such as individual characteristics (e.g., locus of control), trust in robots, self-confidence, task difficulty, task criticality, human abilities, and the capabilities of the robot or automation system. Autonomy levels span from manual control to semi-autonomous and fully autonomous systems.

   In \cite{pasparakis2021control}, the authors conducted a user study in a warehouse order-picking scenario, examining two cases: 1) participants leading the robot and 2) participants following the robot. The results showed that the former approach led to higher productivity, while the latter resulted in improved accuracy. However, findings from “6 River Systems,” a company actively involved in warehouse automation, suggest that granting associates control over their pace results in decreased speed and efficiency \cite{6river}.
            
     The results of a user study involving a household robot showed that participants felt a greater sense of control when the robot operated in semi-autonomous mode. Participants also preferred reduced autonomy for critical tasks, such as scheduling a doctor’s appointment, while they favored greater autonomy for tasks like transporting biscuits from the kitchen to the living room \cite{chanseau2019does}. The user study reported in \cite{schmidbauer2023empirical} indicates that, during a co-assembly task, participants favored maintaining control over decision-making and handling cognitive tasks themselves. In \cite{roy2019automation}, the user study results revealed that participants preferred manual control over automation. The study also emphasized how automation accuracy and task controllability are closely linked, and how these factors significantly influence user satisfaction.

\section{Task Allocation \&  Planning}

\subsection{Problem Statement}\label{sec:2}
This paper considers a collaborative task, $\tau$, consisting of $n$ subtasks and involving two agents: a human and a robot. These agents must cooperate to complete a set of precedence-constrained subtasks, denoted as $\tau = \left\{\tau_1, \tau_2, \dots, \tau_n\right\}$. Each subtask $\tau_i$ has associated completion times, represented as $t_i^h$ for the human and $t_i^r$ for the robot. However, owing to uncertainties stemming from agents and the environment, the task completion time may deviate from the initially specified duration.

In each decision-making step, an agent has the agency to allocate a set of feasible subtasks to the other agent, as well as to assign a subtask to itself and execute it. What sets this problem apart from conventional task allocation and scheduling problems is that the agents here have the autonomy to choose their actions and subtask assignments rather than being assigned specific tasks with predetermined instructions regarding what and when they should execute them. Throughout  this collaboration, the robot needs to:
\begin{itemize}
    \item estimate the human agent's leading/following preference,
    \item  monitor the impact of the human agent's actions on the overall team performance continuously,
    \item  minimize the collaboration cost (e.g., completion time) while adapting to the human agent's preference and performance,
    \item detect and address human errors, if applicable.
\end{itemize}

\subsection{Planning Architecture}
The planning architecture is illustrated in Fig.~\ref{fig:architect}.

\begin{figure*}[!t]
\centering
\scalebox{0.62}{
    \begin{tikzpicture}[%
    ,node distance=10mm
    ,>=latex'
    ,every path/.style={->} ,
    mypostaction/.style n args=3{
   decoration={
       text align={
           left indent=#1},
       text along path, 
       text={|\large| #2},#3
   },
   decorate,
   }
    ]
    \tikzstyle{arrow1} = [-{Latex[length=2mm,width=2mm]}, very thick];
    \tikzstyle{block1} = [draw ,minimum height=10mm,minimum width=28mm ,align=center, rounded corners=.15cm, font=\large, very thick];
    \tikzstyle{output} = [coordinate];
    \definecolor{cl1}{HTML}{F28482};
    \definecolor{cl11}{HTML}{BB0704};
    \definecolor{cl2}{HTML}{F6BD60};
    \definecolor{cl22}{HTML}{BF7600};
    \definecolor{cl3}{HTML}{709DF1};
    
    \node [block1, minimum height = 1.5cm, fill=cl3, draw=blue] (a)  {\Large{State Estimator}};
    \node [block1, right= 1.3cm of a, fill=cl3, draw=blue] (b) {\Large{Robot Planner}};
    \node [block1, right=of b, fill=cl1, draw=cl11] (c) {\Large{Robot}};
    \node [block1, above=of c, yshift=1.5cm, fill=cl1, draw=cl11] (d) {\Large{Human}};
    \node [block1, minimum width=5cm, right=of c, xshift=0.5cm, yshift=-0.75cm, rotate= 90, fill=cl2, draw=cl22] (e) {\Large{Tasks \& Environment}};
    \node [output, right=of e, xshift=0.7cm, yshift=-2.5cm] (o) {};
    \node [inner sep=0,minimum size=0,right of=b, xshift= 0.85cm, label={[label distance=-1.2cm, align=center]90:{\large $\bold S$\\\textbf{\large(Schedule)}}}] (p) {}; 
    \node [inner sep=0,minimum size=0,right of=d, xshift= 0.85cm] (u) {}; 
    \node [inner sep=0,minimum size=0,right of=e] (k) {}; 
    \draw [arrow1] (a) to node[above] {\large \textbf{Belief}} (b);
    \draw [arrow1] (b) to (c);
    \draw [arrow1] (c.east) to node[above] {\large{$\bold {a^R}$}} ($(e.north)+(0,-1.75)$) ;
    \draw [arrow1] (d.east) to ($(e.north)+(0,1.75)$);
    \draw [arrow1] (k) -- +(0,-3.2) -- ($(a.west)+(-0.5,-1.5)$) |- ($(a.west)+(0,-0.5)$);
    \draw [arrow1] (e.south) -- (o.west);
    \draw [arrow1] ($(b.west)+(-0.35,-1.48)$) |- ($(b.west)+(0,-0.3)$);
    \draw [arrow1] (p) -- +(0,1) -- ($(a.west)+(-0.3,1)$) |- ($(a.west)+(0,0.5)$);
    \draw [arrow1] (u) -- +(0,-1) -- node[above] {\large \textbf{Human actions}} ($(a.west)+(-0.5,2.5)$) |- ($(a.west)+(0,0.0)$);
    \node [block1, right = 2cm of o,minimum width=10cm, minimum height = 7.5cm, fill=cl3, draw=blue] (pl) {};
    \node [block1, above = 0.cm of pl.center,minimum width=5cm, minimum height = 2.5cm, draw=black, fill=white] (so) {\scalebox{0.7}{$\begin{aligned}&\mathbf{X}^* = \min_{\left\{\mathbf{X}\right\}}\max_{A} \mathbb{E} \Bigg[\sum_{\tau_i\in \tau, a \in A}X_{\tau_i}^a C_{\tau_i}\big(  a\big)\Bigg] \\
    &\qquad \; \text{subject to}\\
    &\qquad \; \sum_{a\in A}{X_{\tau_i}^a}=1, \quad \forall \,\tau_i\in \tau \\
    &\qquad \; \mathbf{X} \notin F \\
    &\qquad \; \textit{problem-dependent constraints}
    \end{aligned}$}};
    \node [block1, below = 0.7cm of pl.center,minimum width=5cm, minimum height = 2.5cm, draw=black, fill=white] (do) {\scalebox{0.7}{
        $\begin{aligned}
            &\min\max_{\tau_i \in \tau_{\textit{new}}} f_{\tau_i}\\ 
            &\quad\text{subject to}\\
            &\qquad \; P\left(\tau_{i},\tau_{j}\right).f_{\tau_{i}}\leq s_{\tau_{j}}, \qquad  \forall \tau_{i},\tau_{j}\in \tau_{new}\\
            &\qquad \; Q\left(\tau_{i},\tau_{j}\right).f_{\tau_{i}}\leq s_{\tau_{j}}, \qquad  \forall \tau_{i},\tau_{j} \in \tau_{new}\\
            &\qquad \; f_{\tau_i} = s_{\tau_i} + d_{\tau_i}, \qquad \forall \tau_i\in \tau_{new} \\
            &\qquad \; \textit{problem-dependent constraints}
        \end{aligned}$
    }};
     \draw [arrow1] ($(so.west)+(-1.8,-3.7)$) |- ($(so.west)+(0,-0.5)$);
     \draw [arrow1] ($(so.west)+(-4,0.5)$) node[above, xshift=0.6cm] {\large{Belief}} -- ($(so.west)+(0,0.5)$);
     \draw [arrow1] ($(do.west)+(-4,0)$) |- node[above, xshift=0.5cm] {\large{Task State}} ($(do.west)$);
     \draw [arrow1] ($(so.south)$) -- ($(do.north)$);
     \draw[arrow1] (do.east) -| ($(pl.east)+(-2, 0)$) -- ($(pl.east)+(2, 0)$) node[above, xshift=-0.4cm] {S};
     \draw[arrow1] ($(pl.east)+(-2, 0)$) -- ($(pl.east)+(-2, 3.6)$) -| (so.north);
    \node [draw=gray, line width=0.5mm, circle, dashed, below=0mm of b.center, anchor=center, minimum size=32mm] (zp){};
     \path[dashed, gray, line width=0.5mm](zp.south) edge[-] (pl.south west){};
     \path (zp.north) edge[-, dashed, gray, line width=0.5mm] (pl.north west){};

     \draw[-{Latex[bend,length=5em]}, blue!20!white, line width=5ex,postaction={mypostaction={0.01em}{Task selection}{raise=-0.7ex}}] (so)++ (-3,3)  to[bend left]  ($(so.north)+ (-0.5,0)$);
      \draw[-{Latex[bend,length=5em]}, blue!20!white, line width=5ex,postaction={mypostaction={0.01em}{Task scheduling}{raise=-0.7ex}}] (do)++ (-3,-3)  to[bend right]  ($(do.south)+ (-0.5,0)$);
    \end{tikzpicture} }
\caption{Task selection and planning architecture includes the state estimator and robot planner. The robot planner considers human preferences, performance, and task state, followed by two planning steps: task selection and task scheduling. }
\label{fig:architect}
\end{figure*}
\textbf{Tasks \& Environment:} the overall system or task the human and robot collaborate on.

\textbf{Human/Robot:} These two blocks represent the input provided by the human and robot and applied to the system. It's important to note that this is an asynchronous decision-making process, where the human and robot agents act and make decisions independently and at different times. 

\textbf{State estimator}: 
During the collaborative process, the robot evaluates the human's actions and infers their inclination towards leading or following. Furthermore, it is responsible for monitoring the human's performance and assessing their level of performance.  These states, however, cannot be measured directly, and the robot needs to infer them through the history of the interaction. To do so, the state observer takes the history of the human's actions, the robot's beliefs and schedule, and the task state.

\textbf{Robot Planner:} The robot planner is responsible for providing the robot with a schedule based on the task and environment state and the output of the state estimator block, belief about the human agent's preference and performance. The robot planner consists of two phases: task selection and task scheduling. In each decision step, when necessary, the robot performs task selection and subsequently performs task scheduling to determine its following action.

\subsection{Planning Strategy}\label{sec:3}
At each decision step, the robot planner must determine a one-to-one subtask assignment for the agents and establish a task execution schedule to minimize the collaboration cost, injecting both the human agent's preference and performance. 
Task allocation and scheduling problems can usually be modeled as mixed linear integer programs (MILP). However, the complexity of MILP-based solutions makes them computationally intractable. In addition, involving the robot's belief about the human agent's preference and performance adds more complexity to the problem due to the dynamic and unpredictable nature of human behavior and intentions. These challenges, in concert, make formulating and solving the problem as a single optimization problem increasingly demanding and arduous.   

The decomposition method, which divides task planning into task allocation and scheduling, is widely used in the literature to simplify complex planning scenarios. In general, task allocation assigns tasks to specific agents, while sequencing ensures tasks are ordered for execution feasibly and efficiently. Prior works, such as \cite{li2009scheduling} and \cite{ren2009improved}, have shown the effectiveness of this approach in project and job-shop scheduling problems, particularly in enhancing computational efficiency and scalability. These studies typically address task assignments using mixed-integer linear programming (MILP) and solve sequencing through constraint logic programming.

In robotic applications, the decomposition method has proven valuable. In \cite{zhang2016co}, the authors tackled a large-scale multi-agent coordination problem by decomposing it into agent placement, task assignment, and scheduling, using multi-abstraction search to optimize the solution. Similarly, in \cite{qin2024online}, task allocation and sequencing are dynamically adjusted to prevent collisions in multi-robot systems, employing a greedy algorithm to prioritize task execution. In \cite{gombolay2017computational}, the Tercio algorithm is applied in human-robot collaboration, where task allocation is solved first, followed by a fast sequencing subroutine. If the resulting schedule does not meet the desired makespan, the process iterates, finding the next-best task allocation and sequence until a satisfactory solution is achieved.

This paper adopts a similar approach of decomposing the task allocation and sequencing problems. However, we incorporate the uncertainty associated with the human agent’s preferences and performance into the task allocation, framing it as a stochastic optimization problem. The solution from this step is then used to solve a deterministic optimization problem to achieve optimal task sequencing. While this decomposition does not guarantee a joint global optimum across all decision variables, resulting in a suboptimal solution, it allows the robot to develop a viable plan in a timely manner.

In the first step, the robot seeks an optimal task allocation given the set of agents \( A = \{\textit{human}, \textit{robot}\} \) and the set of subtasks \( \tau = \{\tau_1, \tau_2, \dots, \tau_m\} \), where \( m \) represents the number of subtasks that are incomplete or need to be fixed. The function \( C_{\tau_i}(a) \) represents the cost incurred by assigning task \( \tau_i \) to the agent \( a \in A \). This cost considers not only the time required for the agent to complete the subtask but also factors in the human agent’s performance and preferences. In Section~\ref{sec:planning}, Eq.~\ref{eq:cost}, we elaborate on how to implement this function for the specific scenario designed, which can be generalized to various other scenarios with some modifications.

Subsequently, if necessary, a new set of subtasks $\tau_\textit{new} = \tau \cup \tau_\textit{allocate}$ is generated, including actions needed to allocate subtasks to the human  $\tau_\textit{allocate}$. For example, in a sorting task, the robot may place a box on the human agent's side to indicate that the sorting of this box has been assigned to them. This additional subtask, which requires a certain amount of time, is needed as part of the task allocation process. The task scheduler utilizes the derived optimal task allocation and $\tau_\textit{new}$ to determine an optimal task schedule. If the solution achieved in the task allocation phase proves infeasible during the task scheduling phase, the first step must be repeated to obtain a revised allocation.

Here, we provide the general formulation of the task allocation optimization problem, and then in Section \ref{sec:planning}, adapt it to the specific scenario designed.
The task allocation optimization problem can be formulated as \eqref{eq:opt2}, minimizing the maximum cost of task assignments, between the human and robot, 
\begin{align}
    &\mathbf{X}^* = \min_{\left\{\mathbf{X}\right\}}\max_{A} \mathbb{E} \Bigg[\sum_{\tau_i\in \tau, a \in A}X_{\tau_i}^a C_{\tau_i}\big(  a\big)\Bigg] \label{eq:opt2} \\
    &\qquad \; \text{subject to}\nonumber
\end{align}
\vspace{-2em}
\begin{align}
    &\qquad \; \sum_{a\in A}{X_{\tau_i}^a}=1, \quad \forall \,\tau_i\in \tau \label{eq:opt2_cons1}\\
    &\qquad \; \mathbf{X} \notin F \label{eq:opt2_cons3}\\
    &\qquad \; \textit{problem-dependent constraints}.\label{eq:opt2_cons5}
\end{align}
In Eq.~\eqref{eq:opt2}, $\mathbf{X} = \big\{X_{\tau_i}^a  \mid \tau_i \in \tau, a\in A  \big\}$, where $X_{\tau_i}^a \in \big\{0,1\big\}$ is a binary decision variable  that indicates whether task $\tau_i$ is allocated to agent $a\in A$ ($X_{\tau_i}^a = 1$) or not ($X_{\tau_i}^a = 0$). 
Equation \eqref{eq:opt2_cons1} specifies each task must be assigned to exactly one agent, the human or the robot. 
In Constraint~\ref{eq:opt2_cons3}, $F$ represents a set of task allocation solutions that do not result in a feasible task scheduling solution. Consequently, this constraint prohibits the selection of task allocation solutions from this set.
 Eq. \eqref{eq:opt2_cons5} indicates that additional problem-dependent constraints can also be added. An example of a problem-dependent constraint is provided in Eq.~\ref{eq:opt4-const3}, which ensures that there is at least one subtask for the robot to perform at the beginning of each decision step, preventing the robot from remaining idle. After finding a solution for the task allocation problem, we generate $\tau_\textit{new}$ with known required time, $d_{\tau_i}$, to finish each task $\tau_i$ and updated task-precedence constraints.

In the subsequent phase, the robot determines an optimal schedule specifying both the tasks to be performed and their corresponding start and finish times. We introduce decision variables $s_{\tau_i}$, representing the start times of subtask $\tau_i \in \tau_{\textit{new}}$.  $f_{\tau_i}$ denotes the finish time for subtask $\tau_i$. To account for task precedence, we utilize a binary function $P(\tau_{i},\tau_{j})$, which takes the value of 1 when $\tau_i$ must be completed before $\tau_j$. Additionally, we employ a binary decision variable $Q(\tau_{i},\tau_{j})$, where $Q(\tau_{i},\tau_{j})=1$ indicates that both $\tau_i$ and $\tau_j$ are assigned to the same agent, and $\tau_i$ precedes $\tau_j$.

In this paper, considering that we solely focus on collaboration time, the task scheduling problem can be formulated as minimizing the overall processing time, in the following general form:
\begin{equation}
\min\max_{\tau_i \in \tau_{\textit{new}}} f_{\tau_i} \label{eq:opt3} 
\end{equation}   
 \vspace{-2em}
 \begin{align}
     &\text{subject to}\nonumber\\
    &\quad P\left(\tau_{i},\tau_{j}\right).f_{\tau_{i}}\leq s_{\tau_{j}}, &\quad \forall \tau_{i},\tau_{j}\in \tau_{new} \label{eq:opt3_cons2}\\
    &\quad Q\left(\tau_{i},\tau_{j}\right).f_{\tau_{i}}\leq s_{\tau_{j}}, &\quad \forall \tau_{i},\tau_{j} \in \tau_{new}\label{eq:opt3_cons3}\\
    &\quad f_{\tau_i} = s_{\tau_i} + d_{\tau_i}, &\quad  \forall \tau_i\in \tau_{new} \label{eq:opt3_cons4}\\
    &\quad \textit{problem-dependent constraints} \label{eq:opt3_cons5}.
 \end{align}
  Inequality \eqref{eq:opt3_cons2} guarantees the precedence constraints. Inequality \eqref{eq:opt3_cons3} maintains the constraint that each agent can handle only one subtask at a time. In \eqref{eq:opt3_cons4},  $f_{\tau_i}$ is determined by the time needed by the assigned agent, $d_{\tau_i}$, to complete subtask $\tau_i$. It's important to note that the constraints in this optimization problem are specific to the problem being addressed, and as such, the mentioned constraints can be adjusted, and new ones can be introduced \eqref{eq:opt3_cons5}.  An example of this problem-dependent constraint applied to the collaborative scenario in this paper is provided in Eq.~\ref{eq:opt5-const7}. This constraint ensures that the robot always performs a subtask as long as there are unfinished subtasks, thereby avoiding idle states.

Depending on the constraints in optimization problems \eqref{eq:opt2} and \eqref{eq:opt3},  they can be formulated as mixed-integer linear or nonlinear programs. 
Nevertheless, for the collaborative scenario, we will formulate the problem as a mixed-integer linear program and subsequently use solvers such as Gurobi or CPLEX to solve it.

 \subsection{Algorithm}
 The task selection and planning procedure proposed in this paper is explained in Algorithm \ref{alg1}.
 \begin{algorithm}[tb]
 \DontPrintSemicolon
 \caption{Task selection and planning}\label{alg1}
 \Input{Precedence-constrained tasks, $\tau$}
 \SetKwFunction{selection}{TaskSelection} 
 \SetKwFunction{create}{CreatNewTask}
 \SetKwFunction{scheduler}{TaskSchedule}
 \SetKwFunction{action}{GetAction}
 \SetKwFunction{applyaction}{ApplyAction}
$bel \gets$ initial beliefs about human agent's following preference and performance\label{alg1:l1}\;
 \While{ Tasks are not finished}{
    Monitor the human's actions \label{alg1:l3}\;
    Detect the human's errors \label{alg1:l4}\;
    Update task, $\tau$ \label{alg1:l5}\;
    Update $bel$ \label{alg1:l6}\;
    \If{new schedule is needed \label{alg1:l7}}{
        \While{schedule, $S^\star$, is not found \label{alg1:l8}}{
        $X^\star \gets$ \selection{$\tau$, $bel$}\label{alg1:l9}\;
        $\tau_{new} \gets$ \create{$\tau$, $X^\star$}\label{alg1:l10}\;
        $S^\star \gets$ \scheduler{$\tau_{new}$, $X^\star$}\label{alg1:l11}\;
        }\label{alg1:l12}
    }
    $\textit{$a^R$}$ $\gets$ \action{$S^\star$}\label{alg1:l13}\;
    \applyaction{$a^R$}\label{alg1:l14}\;
 }
 \end{algorithm}

 As indicated in Lines \ref{alg1:l1}, the robot initializes its belief about the human agent's performance and following preference.
Throughout the collaboration, until completing all subtasks, the robot monitors and records the human agent's actions and errors (Lines \ref{alg1:l3}-\ref{alg1:l4}). The robot also has to update the task and precedence constraints based on the finished subtasks and the subtasks need to be fixed  (Line \ref{alg1:l5}). Then, based on the planning structure shown in Fig.~\ref{fig:architect}, the robot updates its belief about the human agent's following preference and performance   (Line \ref{alg1:l6}). Next, If the situation requires a new plan, as discussed before, the robot first solves for an optimal task allocation (Line \ref{alg1:l9}) and 
 then creates $\tau_\textit{new}$ (Line \ref{alg1:l10}). Next, the robot solves for an optimal schedule (Line \ref{alg1:l11}). 
If the robot fails to find an optimal schedule with the current task allocation, it proceeds to generate a new task allocation. Subsequently, the robot performs its action, $a^R$ based on the obtained schedule (Lines \ref{alg1:l13}-\ref{alg1:l14}). This continues until the team finishes all subtasks.

The framework outlined in this section has been implemented on a mobile manipulator robot and evaluated in a user study, described below.
\section{User Study: Setup \& Methodology}
This section provides a detailed explanation of the collaborative scenario designed for the study and outlines the study procedure.

\subsection{User Study Setup}

The considerations for designing the user study scenario revolved around three aspects:\begin{enumerate*}
    \item \textbf{Collaboration:} focusing on the human and robot's planning ability, the cobot's better memory, and the human's faster speed,
\item \textbf{Leading/following preference:} the human's agency to adjust their leading/following role,
\item \textbf{Performance:} a task requiring cognitive load and a penalty for mistakes.
\end{enumerate*}

\subsubsection{Setup}

\begin{figure}
    \centering
    \includegraphics[width= \linewidth]{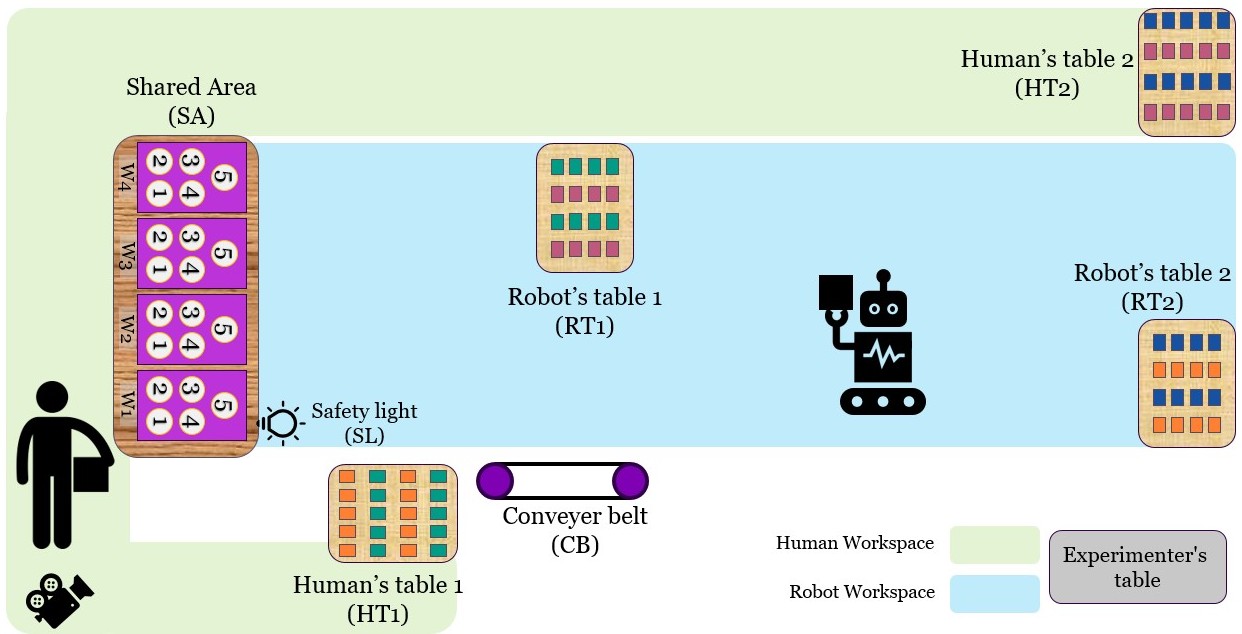}
    \caption{The schematic of the experimental setup}
    \label{fig:exp_env}
\end{figure}

\begin{figure}
    \centering
    \includegraphics[ width=\linewidth, height=0.6\linewidth]{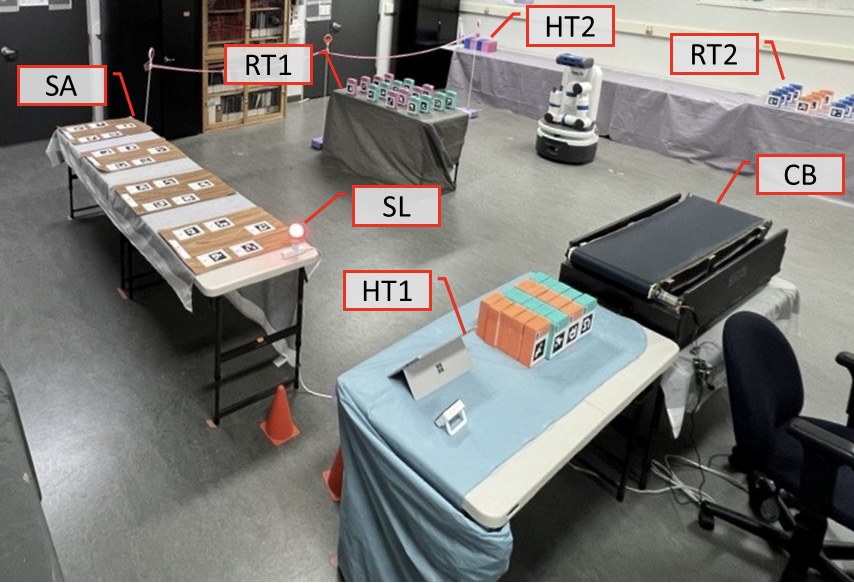}
    \caption{A view of the experiment environment, taken from the location of the camera in Fig.~\ref{fig:exp_env}. The Fetch robot is positioned in its workspace between its two designated tables (RT1 and RT2), as illustrated in Fig.~\ref{fig:exp_env}. The nearest table in the figure, located next to the conveyor belt, belongs to the human (HT1). On this table, there is also a tablet on which the GUI is installed. The human agent's other table is situated at the corner of the room (HT2). Four workspaces are present on the shared area (SA), along with a light bulb (SL). Safety tapes and cones separate the work areas of the agents.}
    \label{fig:exp_env_real}
\end{figure}

Fig.~\ref{fig:exp_env} and Fig~\ref{fig:exp_env_real} illustrate the experimental setup. The location of the camera in Fig.~\ref{fig:exp_env} shows the location where the picture in Fig.~\ref{fig:exp_env_real} was taken.
Both the human and the robot work in designated areas, separated by safety tape and cones.  The collaborative task involves arranging colored blocks on four workspaces, namely W\textsubscript{1} to W\textsubscript{4}, within a shared area (table). 
Each workspace comprises five numbered spots, and participants must adhere to the numerical order when placing blocks. For instance, in W\textsubscript{2}, they must fill spot 1 before proceeding to spot 2. While switching between workspaces is allowed, the spot order must be strictly followed.

Essentially, blue blocks are distant from both agents, while green blocks are close to both. Orange blocks are far from the robot but close to the human agent, while pink blocks are close to the robot and distant from the human agent. Table 1 provides a summary of the block distribution.

    \begin{table}
    \renewcommand{\arraystretch}{1.2}
    \centering
        \caption{Distribution of blocks with respect to the distance to the shared area}
    \label{tab:color_dist}
    \begin{tabular}{@{}lll}
    \toprule
        \textbf{Color} & \textbf{Human} & \textbf{Robot} \\ \midrule
        Green  & Close (HT1) & Close (RT1)\\
        Pink  & Close (HT1) & Far (RT2)\\
        Orange & Far (HT2) & Close (RT1) \\
        Blue & Far (HT2)& Far (RT2)\\
         \bottomrule
    \end{tabular}
\end{table}

\begin{figure} 
    \centering
  \subfloat[Pattern A\textsubscript{1} \label{fig:pattern_a1}]{%
       \includegraphics[width=0.45\linewidth]{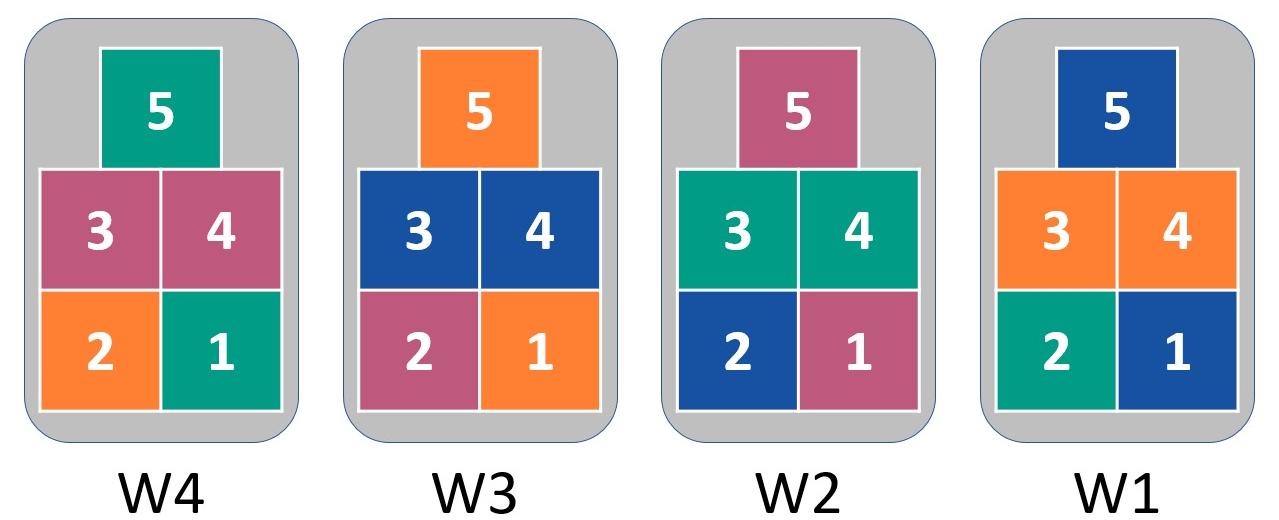}} 
    \hfill
  \subfloat[Pattern A\textsubscript{2} ]{%
        \includegraphics[width=0.45\linewidth]{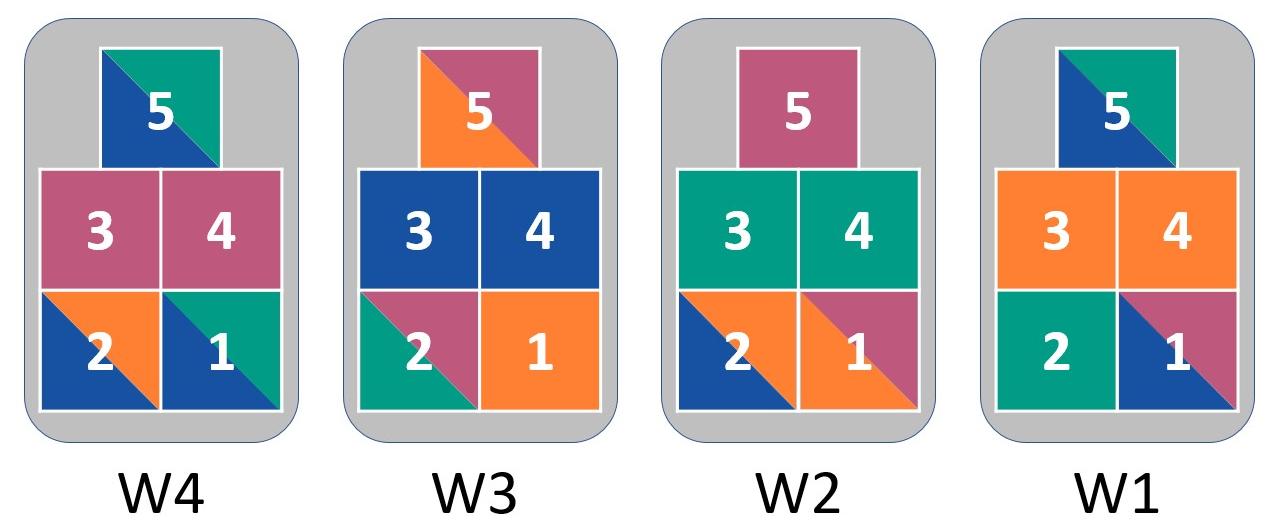}}
    \\
  \subfloat[Pattern B\textsubscript{1} ]{%
        \includegraphics[width=0.45\linewidth]{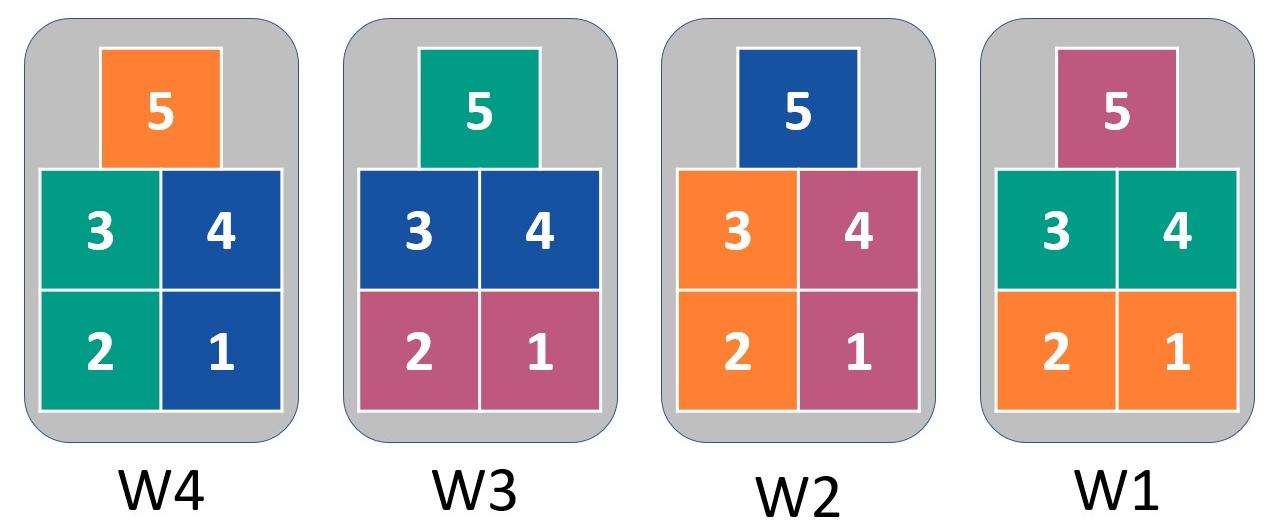}}
    \hfill
  \subfloat[Pattern B\textsubscript{2} ]{%
        \includegraphics[width=0.45\linewidth]{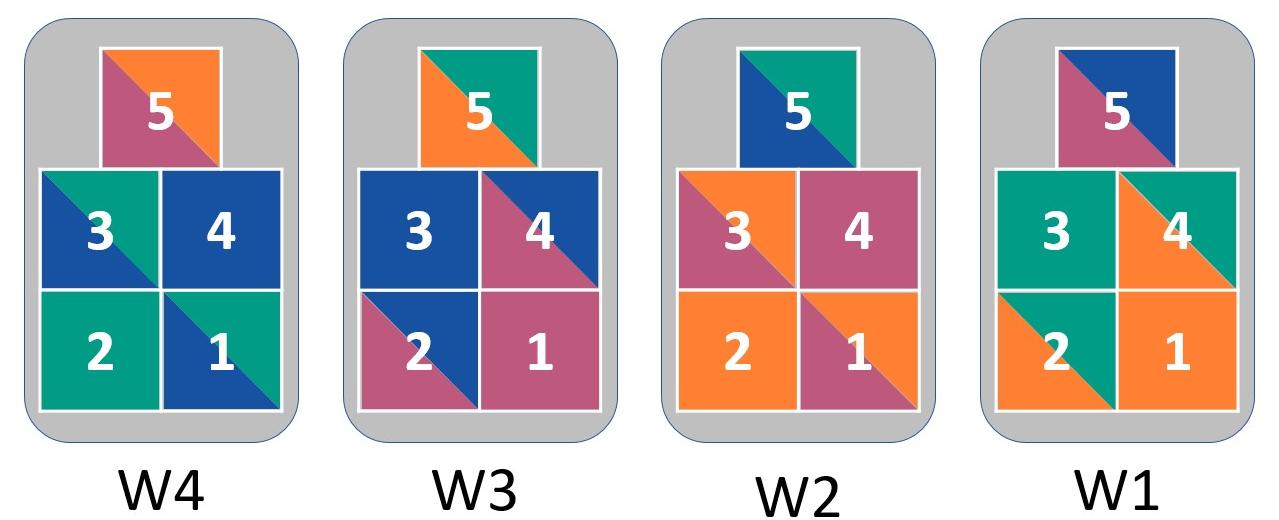}}
    \\
  \subfloat[Pattern C\textsubscript{1} ]{%
        \includegraphics[width=0.45\linewidth]{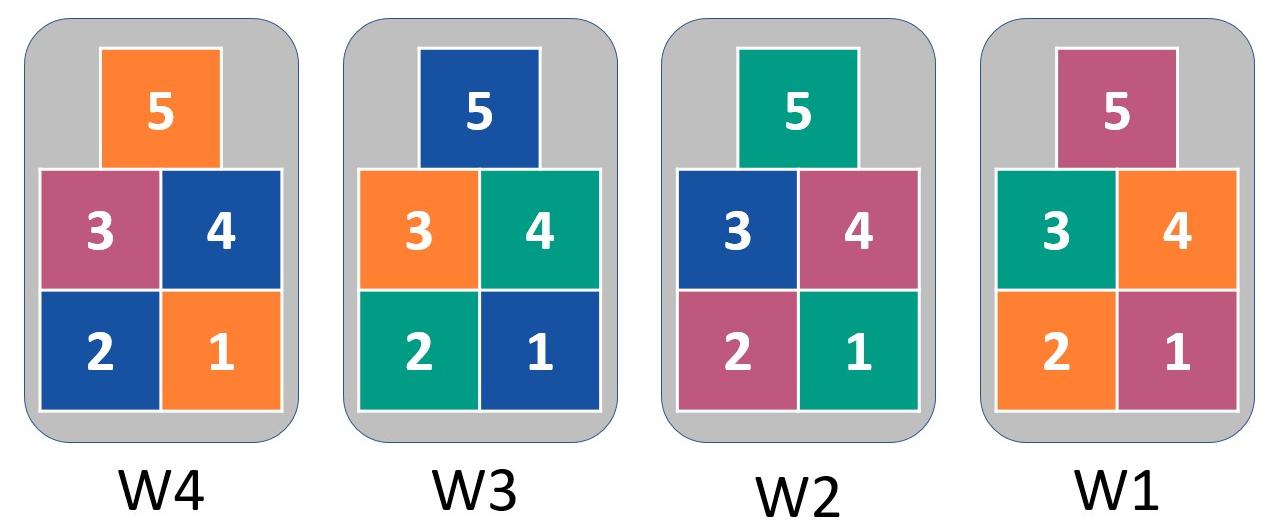}}
    \hfill
  \subfloat[Pattern C\textsubscript{2} ]{%
        \includegraphics[width=0.45\linewidth]{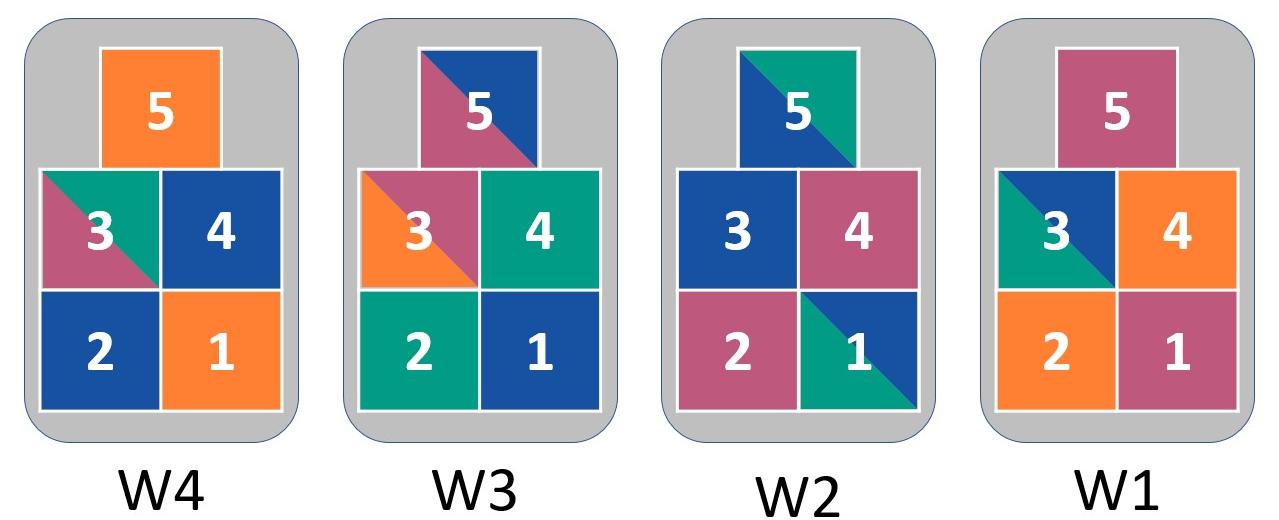}}
        \\
  \subfloat[Pattern D\textsubscript{1} ]{%
        \includegraphics[width=0.45\linewidth]{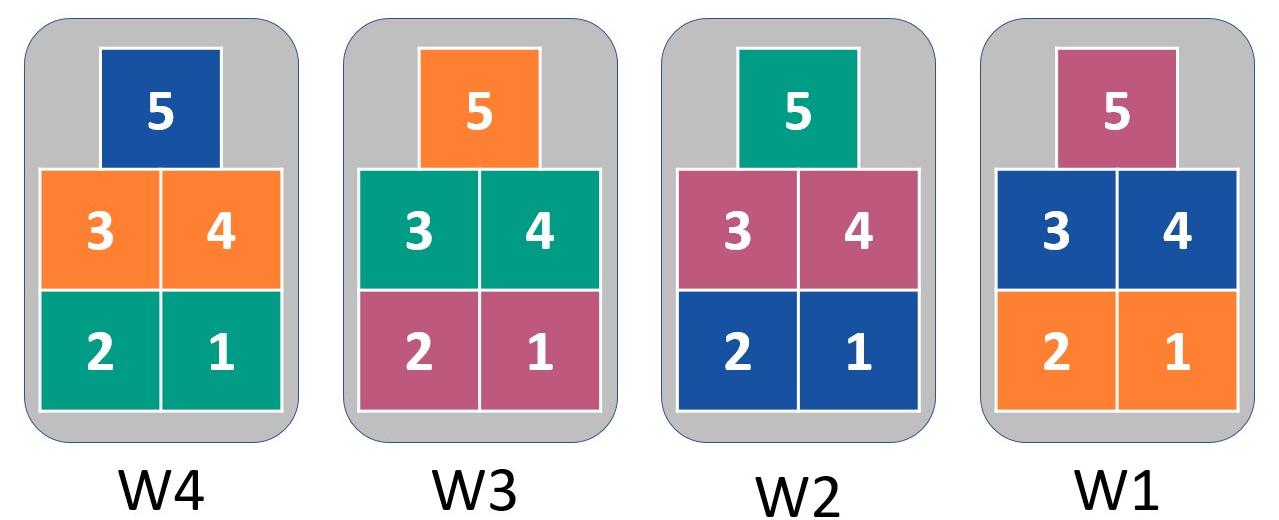}}
    \hfill
  \subfloat[Pattern D\textsubscript{2} ]{%
        \includegraphics[width=0.45\linewidth]{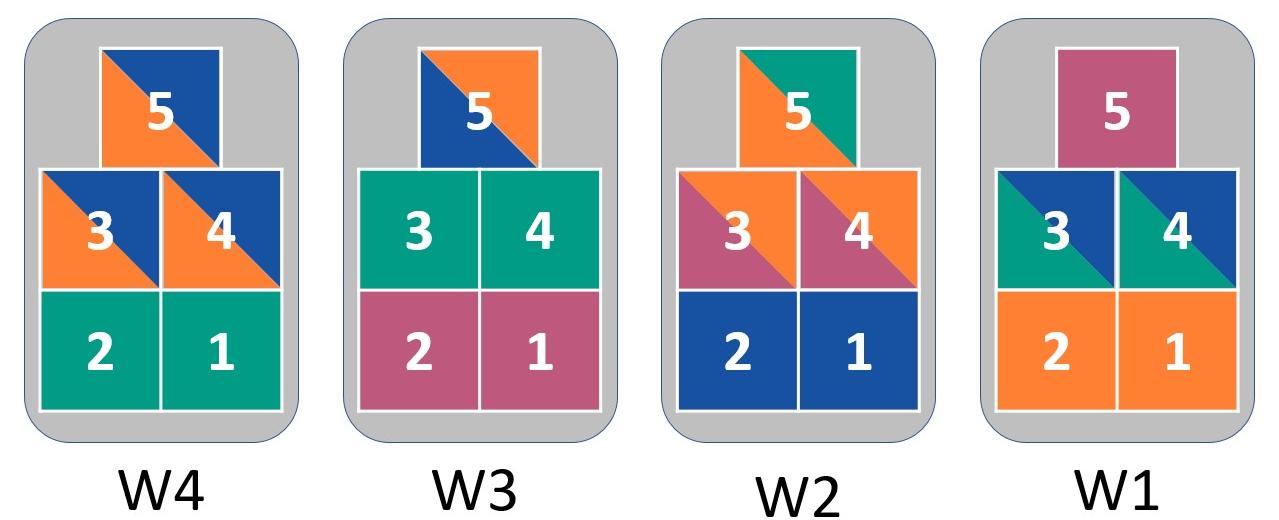}}
  \caption{\textbf{a, c, e, g:} Patterns A\textsubscript{1}, B\textsubscript{1}, C\textsubscript{1} and D\textsubscript{1} are the patterns, printed on the sheets of paper, that participants have to memorize in 45 seconds and then return it to the experimenter. \textbf{b, d, f, h:} Patterns A\textsubscript{2}, B\textsubscript{2}, C\textsubscript{2} and D\textsubscript{2} are the patterns with partially unknown spot, and participants can keep it until the end of the collaborative task as a hint to recall the first pattern}
  \label{fig:patterns} 
\end{figure}

To complete the workspaces (spots), the team must adhere to a prescribed pattern of colors. The pattern shown in Fig.~\ref{fig:pattern_a1} represents a sample pattern the human-robot team is expected to follow when filling the spots. The structure of these patterns precisely mirrors that of the workspaces within the shared area. Initially, participants are presented with a fully known version of the pattern (Patterns A\textsubscript{1}, B\textsubscript{1}, C\textsubscript{1}, and D\textsubscript{1}), such as A\textsubscript{1}, printed on a sheet of paper. They are given 45 seconds to memorize it and then return it to the experimenter. Subsequently, the experimenter provides them with a partially known version of the same pattern (Patterns A\textsubscript{2}, B\textsubscript{2}, C\textsubscript{2}, and D\textsubscript{2}), for example, A\textsubscript{2}. In this partially known version, certain spots contain two colors, with only one of them being correct. Essentially, the partially known pattern acts as a cue, helping participants recall the initially presented pattern. Participants are allowed to retain the partially known version throughout the duration of the task.

The robot needs to return the block to the human agent via the conveyor belt (see Fig.~\ref{fig:exp_env}) in cases the human places the wrong block on the shared table. Additionally, a red light bulb in the shared area warns the human agent not to place or remove any blocks from the table as the robot approaches. Participants, however, can continue planning, moving into their work area, and picking up blocks from other tables. This, in addition to safety concerns, helps control the human agent's speed and prevents them from moving fast and perceiving the collaboration as a race.

For this study, we employed the Fetch mobile manipulator robot \cite{wise2016fetch} and programmed it for autonomous navigation within its designated work area, performing the pick-and-place task (see Appendix\ref{append}). However, 
For safety, the experimenter closely monitors the robot and can take control with a joystick or stop it using the emergency button if needed. The following details provide additional information about the study setup:

\begin{itemize}[leftmargin=*]
    \item The collaborative scenario was inspired by the kitting task. The kitting task involves gathering a specific set of components for a defined purpose, which is then directed to workstations for the assembly of intermediate or end products \cite{tung2022bilevel}.
    \item We chose the pattern memorization task to present participants with a cognitive challenge within the limitations of a brief collaboration scenario. Conducting a lengthy experiment that might mentally and physically strain participants would have been impractical and difficult to obtain ethics approval. Consequently, we selected relatively concise tasks, each lasting around 12-20 minutes, while ensuring they maintain mental engagement.
    
    \item  Patterns A\textsubscript{2}, B\textsubscript{2}, C\textsubscript{2}, and D\textsubscript{2} feature 9, 12, 6, and 9 partially known spots, respectively. This intentional variation in the number of partially known spots aims to introduce distinct difficulty levels and cognitive load.

    \item Participants were instructed to handle only one block at a time, aligning with the cobot's gripper capacity to grasp a single block.

    \item The distribution of blocks includes ten of each color on the human's tables, while the robot's tables accommodate eight of each color. With eighteen blocks of each color in total, exceeding the required five per pattern, this distribution accounts for potential mistakes. More blocks were placed on the human's table due to their faster working pace.

    \item Each block is equipped with an ArUco marker, facilitating the robot in locating and grasping blocks.

\end{itemize}

\subsubsection{Tasks} Each participant is asked to complete four tasks:
\begin{itemize}[leftmargin=*]
    \item Task 0: In this task, participants work alone, without the robot. For all participants, we use pattern A and follow the same procedure.
    This task, in addition to being a practice for participants to learn how to do the task and place the blocks, provides useful information regarding the participants' performance, self-confidence, and perceived workload, which its details and results are beyond the scope of the present manuscript and are detailed in \cite{fetch_human}.

    \item Tasks 1, 2, 3: The robot joins the human in these tasks. We consider six different permutations (modes) of the order of patterns B, C, and D (e.g., \{B, C, D\}, \{C, B, D\}) and randomly assign participants to one of these six modes in a way that has a balance across the modes. Following the same procedure, we asked participants to memorize the first pattern in 45 seconds and then return it. Next, we provide them with the second pattern. In addition, to resemble real cobot scenarios where mistakes have a cost associated, we informed participants that for each misplaced block on the table, at the end of the task, when they declare finishing the task, 1\$ would be deducted from the total remuneration amount. Providing this disinformation, considered a type of deception, was approved by the University of Waterloo Human Research Ethics Board.
\end{itemize}

\subsubsection{Agents' Actions}
We considered a set of six different actions for each agent, the human and the robot. These actions, listed in Table~\ref{tab:actions}, are the same for both agents, providing them with a similar level of agency. Note that the feasibility of actions depends on the state of the task, and at each decision step, some of them may not be feasible. For example, when no task has been allocated to the robot, action H4, performing a task assigned by the human, is not applicable. 

"Leading" actions generally involve task assignment to the other agent (H2 and R2) or rejecting tasks assigned by the other agent (H6 and R6). "Following" actions primarily involve executing tasks assigned by the other agent (H4 and R4). For the human, not performing action H1, depending on the task state, can indicate a preference for following, such as when the human forgets the pattern and waits for the robot's help. Additionally, actions H1, H2, H3, and H6 can be used to evaluate the human's performance. Action H4 is inherently correct, as the robot always assigns the correct subtask and color. The robot also considers human action H5 (canceling a wrong assignment) to update its estimate of the human's preferences and performance. Action H1, depending on the selected subtasks and task state, may be perceived as a more neutral action by the human. For example, if the human's accuracy is high and they choose blue blocks (which are located far from both agents), maintaining high team performance, the robot may not need to assign additional subtasks to the human and can allow the human to continue with their plan.

To give the robot more autonomy in adjusting or reassuming control when human performance is poor, we made the task rejection option (Action H6) require extra effort. It involves performing H4 (task execution) and H3 (returning the task) on the GUI without physical action. Fig.~\ref{fig:stategraph} displays the state graph of a single subtask, considering the possible actions of both human and robot agents as outlined in Table~\ref{tab:actions}.

\begin{table*}
    \centering
      \caption{The human and robot's sets of actions}
    \label{tab:actions}
    \begin{tabular}{@{}l|l@{}}
    \toprule
        \multicolumn{1}{c}{\textbf{Human}} & \multicolumn{1}{c}{\textbf{Robot}} \\ \midrule
        H1- Selecting a task for themselves  & R1- Selecting a task for itself \\
        H2- Assigning a task to the robot  & R2- Assigning a task to the human\\
        H3- Returning a block from the shared workspace & R3- Returning a \textit{wrong} block from the shared workspace \\
        H4- Performing a task assigned by the robot & R4- Performing a \textit{correct} task assigned by the human \\
        H5- Canceling a task assigned to the robot & R5- Canceling a task assigned to the human \\
        H6- Rejecting a task assigned by the robot  & R6- Rejecting a task assigned by the human\\ \bottomrule
    \end{tabular}
\end{table*}
\begin{figure}[t]
    \centering
    \scalebox{0.4}{
    \begin {tikzpicture}[-latex ,auto ,node distance =6 cm and 2cm ,on grid ,
semithick]
\tikzstyle{state} =[circle ,color =white, draw, black , text=black , minimum width =4.0 cm, font=\bfseries\Large, align=center]
\tikzstyle{arrow1} = [font=\bfseries\Large, {Latex[length=2mm,width=2mm]}-]
\tikzstyle{arrow2} = [font=\bfseries\Large, -{Latex[length=2mm,width=2mm]}]

\node[state] (s0) {Initial State};
\node[state] (s3) [right=8cm of s0] {Placed correctly};
\node[state] (s2) [below right = 6cm and 5cm of s0] {Assigned to \\robot correctly};
\node[state] (s1) [below left =6cm and 6cm of s0] {Assigned to \\robot incorrectly};
\node[state] (s4) [above right = 6cm and 5cm of s0] {Assigned to \\human};
\node[state] (s5) [above left =6cm and 6cm of s0] {Misplaced};

\path[arrow1] (s0) edge [bend left =20]  node[sloped, below] {H3} (s5);
\path[arrow1] (s0) edge [bend right =20]  node[sloped, above] {R3} (s5);
\path[arrow2] (s0) edge  node[sloped, above] {H1} (s5);

\path[arrow1] (s0) edge [bend left =20]  node[sloped, below] {H5} (s1);
\path[arrow1] (s0) edge [bend right =20]  node[sloped, above] {R6} (s1);
\path[arrow2] (s0) edge  node[sloped, below] {H2} (s1);

\path[arrow2] (s0) edge [bend left =20]  node[sloped, above] {H1} (s3);
\path[arrow2] (s0) edge [bend right =20]  node[sloped, below] {R1} (s3);
\path[arrow1] (s0) edge  node[sloped, above] {H3} (s3);

\path[arrow1] (s0) edge [bend left =20]  node[sloped, above] {R5} (s4);
\path[arrow1] (s0) edge [bend right =20]  node[sloped, below] {H6} (s4);
\path[arrow2] (s0) edge  node[sloped, above] {R2} (s4);

\path[arrow1] (s0) edge [bend left =10]  node[sloped, above] {H5} (s2);
\path[arrow2] (s0) edge [bend right =10]  node[sloped, below] {H2} (s2);

\path[arrow2] (s4) edge   node[sloped, above] {H4} (s3);
\path[arrow2] (s2) edge   node[sloped, below] {R4} (s3);
\end{tikzpicture}
}
\caption{State graph of a single subtask. There are six potential states for each subtask based on the actions detailed in Table 1. Initially, each subtask is in the ``Initial state" state. To complete a subtask, it must transition to the ``Placed correctly" state and remain in this state. The states ``Assigned to robot correctly" and ``Assigned to robot incorrectly" occur when the human agent respectively assigns a subtask to the robot with the correct or wrong color. The ``Misplaced" state is reached when the human places a wrong color on the shared area for a subtask. The state ``Assigned to human" happens when the robot assigns a subtask to the human agent.}
\label{fig:stategraph}
\end{figure}

\subsubsection{Human-robot communication} 
\begin{figure}
    \centering
    \includegraphics[width=\linewidth]{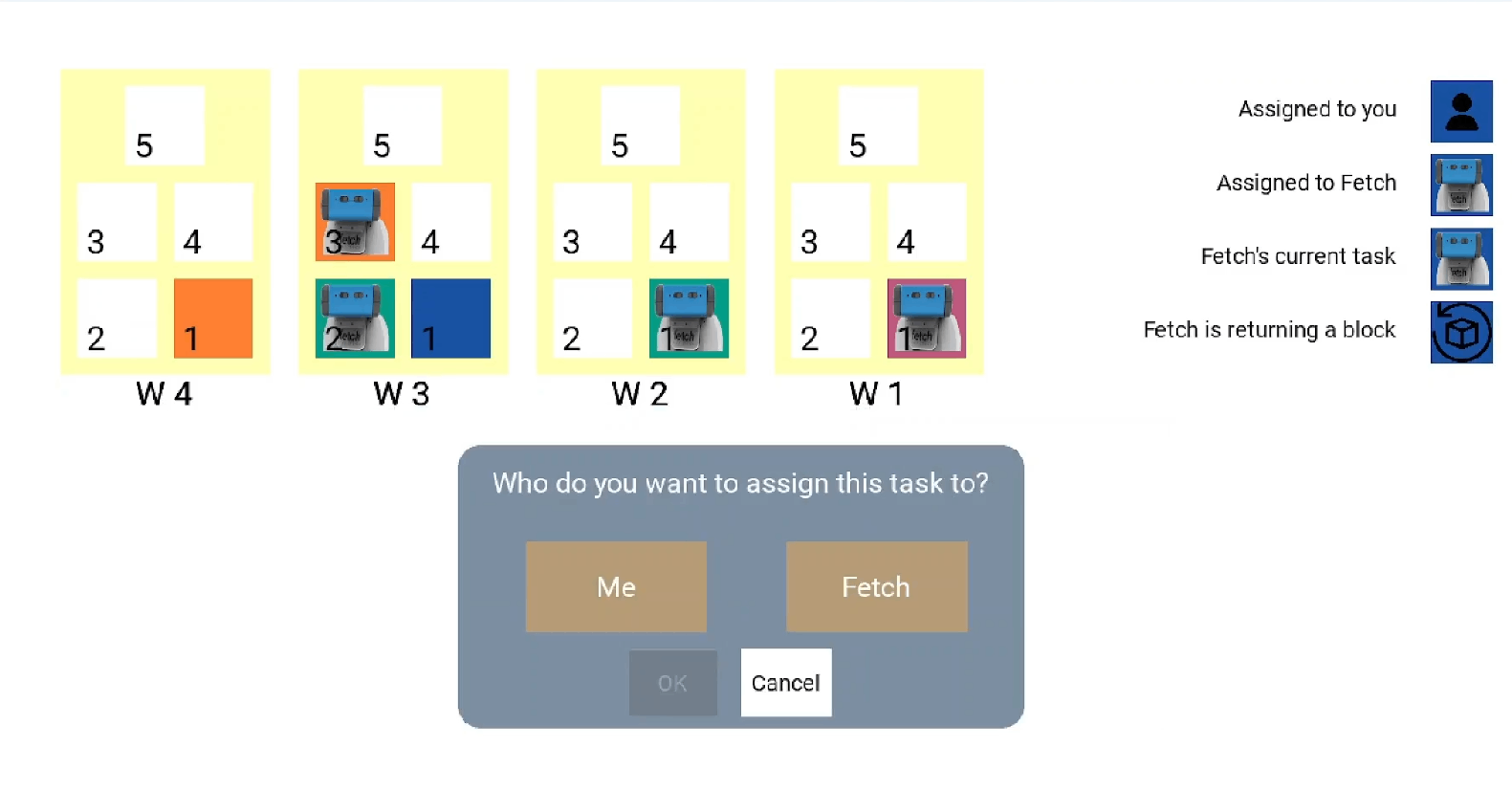}
    \caption{A screenshot of the GUI through which participants can communicate their actions to the robot and receive information about the robot's decisions and actions (actions listed in Table~\ref{tab:actions})}
    \label{fig:gui}
\end{figure}

Both agents need to communicate to inform each other about their next actions (see Table~\ref{tab:actions}). This can be done through a graphical user interface (GUI) designed and installed on a tablet. Participants could leave it on a table in the room or hold it. This helps them assign tasks to the robot and inform it about their next action. Similarly, the robot can assign tasks to the participant via the GUI and inform them about its actions. Fig.~\ref{fig:gui} shows a screenshot of the GUI. The GUI restricts the human agent from taking unfeasible actions, such as violating precedence constraints or choosing tasks already underway by the robot. We instructed participants how to work with the GUI and let them try it once before starting the tasks. Participants were also asked to scan the marker on the blocks before placing them on the shared area, as the robot has to know the block's ID if it needs to return it. This is implemented on the GUI, automatically launching the tablet's camera and letting participants scan the marker. We avoid the details of the GUI's design and implementation for the sake of brevity.
\begin{remark}
    If participants need to reject an assigned task by the robot, they first have to do the assigned task (H4) and then execute the returning action (H3), performing both actions on the GUI, without needing to do them physically.
\end{remark}

\subsection{Adaptation \& Planning} \label{sec:planning}
Here, we explain the components and details of the task planning architecture implemented for the designed scenario.  

\textbf{Collaborative task}: 
The tasks and their associated precedence constraints are conveyed using a directed acyclic graph termed the task graph. This graph depicts tasks as vertices, while edges signify their precedence constraints. Fig.~\ref{fig:taskgraph} provides a visual representation of the initial task graph for the experiment, including two dummy nodes designating the starting ($\tau_0$) and finishing ($\tau_{21}$) points.

In this experiment, we assume the same and time-invariant speed for all participants, the average human walking speed (1.2 $m/s$). In addition, we tested the required time for the robot to complete the pick-and-place scenario from different tables and used the average time for the near and far tables. Thus, the nominal required processing times for tasks are fixed for both the human and the robot.
\begin{figure}[t]
    \centering
    \scalebox{0.58}{
    \begin {tikzpicture}[-latex ,auto ,node distance =1.5 cm and 2cm ,on grid ,
semithick]
\tikzstyle{state} =[circle ,color =white, draw, black , text=black , minimum width =1.0 cm, font=\boldmath\Large]
\tikzstyle{arrow1} = [-{Latex[length=2mm,width=2mm]}]
\tikzstyle{arrow2} = [-{Latex[length=2mm,width=2mm]}, dashed]
\node[state] (t1) {$\tau_1$};
\node[state] (t0) [below left= 2.25cm and 2cm of t1] {$\tau_0$};
\node[state] (t6) [below =of t1] {$\tau_6$};
\node[state] (t11) [below =of t6] {$\tau_{11}$};
\node[state] (t16) [below =of t11] {$\tau_{16}$};
\path[arrow2] (t0) edge [bend left =20]  (t1);
\path[arrow2] (t0) edge [bend left =20]  (t6);
\path[arrow2] (t0) edge [bend right =20]  (t11);
\path[arrow2] (t0) edge [bend right =20]  (t16);

\node[state] (t2) [right =of t1] {$\tau_2$};
\path[arrow1] (t1) edge (t2);
\node[state] (t3) [right =of t2] {$\tau_3$};
\path[arrow1] (t2) edge (t3);
\node[state] (t4) [right =of t3] {$\tau_4$};
\path[arrow1] (t3) edge (t4);
\node[state] (t5) [right =of t4] {$\tau_5$};
\path[arrow1] (t4) edge (t5);

\node[state] (t7) [right =of t6] {$\tau_7$};
\path[arrow1] (t6) edge (t7);
\node[state] (t8) [right =of t7] {$\tau_8$};
\path[arrow1] (t7) edge (t8);
\node[state] (t9) [right =of t8] {$\tau_9$};
\path[arrow1] (t8) edge (t9);
\node[state] (t10) [right =of t9] {$\tau_{10}$};
\path[arrow1] (t9) edge (t10);

\node[state] (t12) [right =of t11] {$\tau_{12}$};
\path[arrow1] (t11) edge (t12);
\node[state] (t13) [right =of t12] {$\tau_{13}$};
\path[arrow1] (t12) edge (t13);
\node[state] (t14) [right =of t13] {$\tau_{14}$};
\path[arrow1] (t13) edge (t14);
\node[state] (t15) [right =of t14] {$\tau_{15}$};
\path[arrow1] (t14) edge (t15);

\node[state] (t17) [right =of t16] {$\tau_{17}$};
\path[arrow1] (t16) edge (t17);
\node[state] (t18) [right =of t17] {$\tau_{18}$};
\path[arrow1] (t17) edge (t18);
\node[state] (t19) [right =of t18] {$\tau_{19}$};
\path[arrow1] (t18) edge (t19);
\node[state] (t20) [right =of t19] {$\tau_{20}$};
\path[arrow1] (t19) edge (t20);

\node[state] (t21) [below right = 2.25cm and 2cm of t5] {$\tau_{21}$};
\path[arrow2] (t5) edge [bend left =20]  (t21);
\path[arrow2] (t10) edge [bend left =20]  (t21);
\path[arrow2](t15) edge [bend right =20]  (t21);
\path[arrow2] (t20) edge [bend right =20]  (t21);
\end{tikzpicture}
}
\caption{Task graph of the experiment, including twenty subtasks ($\tau_1$ - $\tau_{20}$) and two dummy nodes representing starting ($\tau_{0}$) and finishing points ($\tau_{21}$).}
\label{fig:taskgraph}
\end{figure}

\textbf{Following preference and performance}: The robot uses the single scalar random variables $\alpha_f$ and $\alpha_e$, respectively, to capture the human agent's following preference and performance (i.e., error-proneness or inaccuracy). For both, we consider a tuple of possible discrete values, $W=(w_0, \dots, w_{10}) = (0, 0.1, 0.2, \dots, 1)$ with steps of $0.1$. In other words, there exists $i\in\{0,\dots,10\}$ such that $w_i = \alpha_f$ (the same for $\alpha_e$).  For $\alpha_f$, a value closer to zero indicates a leading preference, while values closer to one suggest a following preference. Additionally, values of $\alpha_e$ closer to one represent higher error-proneness (low accuracy), while values closer to zero indicate lower error-proneness (higher accuracy). These discrete steps are chosen because they are accurate enough to effectively capture the human's preference and performance.

At the beginning of the session, as a default, the robot assumes that the human agent prefers to follow it and has high accuracy, so the robot sets its initial belief about the human agent as follows:
\begin{align*}
    &P[\alpha_f=w_i] = b(i; n=10, p=0.7), \; i\in\{0,\dots,10\},
\end{align*}
and similarly,
\begin{align*}
    &P[\alpha_e=w_i] = b(i; n=10, p=0.1), \; i\in\{0,\dots,10\},
\end{align*}
where $b(i;n,p)$ is a binomial distribution.

\textbf{Task allocation}: 
 The problem of task allocation is formulated using \eqref{eq:opt2}-\eqref{eq:opt2_cons5}. 
Designing the function $C_{\tau_i}$ is the most critical aspect of this optimization problem, as it determines how the robot selects tasks for itself while considering the task completion time, the human agent's preferences and performance, and the need to avoid frequent plan changes. Furthermore, $C_{\tau_i}$ must impose a higher cost for assigning tasks to a human who prefers to lead, and a lower cost for a human with low performance. In this scenario, having the robot allocate tasks reduces the human's control, thereby mitigating the potential for human errors.

\begin{align}
         C_{\tau_i}\left(a\right)= \begin{cases}
     t_i^h  \alpha_f + c_f(1-\alpha_f) + c_v x_{\tau_i}^{\textit{robot}} \qquad &a=\textit{Human}\\
     t_i^r + \alpha_e c_e + c_v x_{\tau_i}^{\textit{human}} &a=\textit{Robot}
     \end{cases}. \label{eq:cost}
\end{align}
The parameter $c_f$ represents the penalty incurred when assigning the subtask to a human agent who prefers to lead, while $c_e$ denotes the penalty imposed for not allocating subtasks to the human agent who makes mistakes.
Assigning subtasks to the human agent allows the robot to inform them about the next blocks to be placed on the shared table, minimizing the risk of wrong decisions by the human agent.
Additionally, a penalty cost $c_v$ is applied when the robot (human) is assigned a subtask that has already been allocated to the human (robot) agent, $x_{\tau_i}^{\textit{human}}=1$ ($x_{\tau_i}^{\textit{robot}}=1$). This penalty prevents frequent and substantial changes in task allocation. In addition, as a problem-dependent constraint, we need to add a constraint ensuring the allocation of at least one subtask from $\tau$ to the robot at each decision step. This constraint ensures the robot will start {doing} another subtask  or {correcting} the human agent's errors.

The derivation of optimization problem \ref{eq:opt2} for the task within the designed scenario is equivalent to the optimization problem represented by \ref{eq:lin1}, which is formulated as a Mixed-Integer Linear Program (MILP). In this context, we introduce decision variables $q_i\in Q$ for $i\in I =\{1, \dots, n_t\}$ to determine the assignment of agents to subtasks, where $n_t$ represents the total number of remaining tasks. Specifically, when $q_i = 1$, it signifies that subtask $i$ is assigned to the human agent, while $q_i = 0$ indicates the allocation of the subtask to the robot.
\begin{align}\label{eq:lin1}
    Q^* = &\arg\min_{\left\{{q_i}\right\}}z
\end{align}
\vspace{-2em}
\begin{align}
    &\quad \; \text{subject to}\nonumber\\
    &\qquad z - \sum_{i\in I}{q_i\Big(t_i^h  p_f + c_f(1-p_f) + c_hx_{\tau_i}^{\textit{robot}}\Big)} \geq 0\\
    &\qquad z - \sum_{i\in I}{(1-q_i)\Big(t_i^r + p_e c_e + c_r x_{\tau_i}^{\textit{human}}\Big)} \geq 0\\
    &\qquad \sum_{i\in U}{q_i} \leq \left|U\right|-1, \label{eq:opt4-const3}
\end{align}
where $U$ is the set of indexes of all feasible subtasks that the robot can perform immediately by placing the blocks on the shared area. This inequality ensures that the robot will start placing a new block after placing the previous one.

Fig.~\ref{fig:allocatedtaskgraph} illustrates an example of the updated task graph following task allocation. Subtasks 6 and 16 ($\tau_{6}$ and $\tau_{16}$) are shaded in gray to indicate their completion. However, for subtasks 1 and 17, the human agent mistakenly placed the wrong colored blocks on the shared area. Consequently, the robot must rectify these errors by executing subtasks $\tau_1^e$ and $\tau_{17}^e$. Subtask 11 ($\tau_{11}$) has already been assigned to the human. Additionally, the robot has allocated subtasks 1, 2, 4, 5, 7, 9, 10, 13, 18, and 20 to the human. Considering precedence constraints, the robot can initiate subtask $\tau_7^a$ to allocate and instruct the human agent through the GUI to perform subtask 7 ($\tau_7$). The robot has also allocated subtasks 3, 8, 12, 14, 15, 17, and 19 to itself.

\begin{figure}[t]
    \centering
    \scalebox{0.55}{
    \begin {tikzpicture}[-latex ,auto ,node distance =2cm and 2.5cm ,on grid ,
semithick]
\tikzstyle{state} =[circle ,color =white, draw, black , text=black , minimum width =1.1 cm, font=\boldmath\LARGE]
\tikzstyle{human} =[circle , draw, black , fill=BurntOrange!70, text=black , minimum width =1.1 cm, font=\boldmath\LARGE]
\tikzstyle{robot} =[circle , draw, black , fill=Blue!70, text=white , minimum width =1.1 cm, font=\boldmath\LARGE]
\tikzstyle{allocate} =[circle , draw, black , fill=Orchid!70, text=black , minimum width =0.9 cm, font=\boldmath\Large]
\tikzstyle{error} =[circle , draw, black , fill=BrickRed!70, text=white , minimum width =1 cm, font=\boldmath\Large]

\tikzstyle{arrow1} = [-{Latex[length=2mm,width=2mm]}]
\tikzstyle{arrow2} = [-{Latex[length=2mm,width=2mm]}, dashed]
\node[human] (t1) {$\tau_1$};
\node[state] (t0) [fill=black!60, thin, dashed, text=white, below left= 3.cm and 1.5cm of t1] {$\tau_0$};

\node[allocate] (t1a) [below left= 0cm and 1.5cm of t1] {$\tau_1^a$};
\node[error] (t1e) [below left= 1.5cm and 2.9cm of t1] {$\tau_1^e$};
\path[arrow2] (t0) edge[bend left =20] (t1e);
\path[arrow1] (t1e) edge [bend left =20](t1a);
\path[arrow1] (t1a) edge (t1);

\node[state] (t6) [fill=black!30, thin, dashed, text=black, below =of t1] {$\tau_6$};
\node[state] (t11) [fill=green!30, thin, below =of t6] {$\tau_{11}$};
\node[state] (t16) [fill=black!30, thin, dashed, text=black, below =of t11] {$\tau_{16}$};

\path[arrow2] (t0) edge [bend left =20]  (t6);
\path[arrow2] (t0) edge [bend right =20]  (t11);
\path[arrow2] (t0) edge [bend right =20]  (t16);

\node[human] (t2) [right =of t1] {$\tau_2$};
\node[allocate] (t2a) [above right= 1.0cm and 1.25cm of t1] {$\tau_2^a$};
\path[arrow1] (t1) edge  (t2);
\path[arrow1] (t1) edge [bend left=20] (t2a);
\path[arrow1] (t2a) edge [bend left=25] (t2);
\node[robot] (t3) [right =of t2] {$\tau_3$};
\path[arrow1] (t2) edge (t3);
\node[human] (t4) [right =of t3] {$\tau_4$};
\node[allocate] (t4a) [above right= 1.0cm and 1.25cm of t3] {$\tau_4^a$};
\path[arrow1] (t3) edge (t4);
\path[arrow1] (t3) edge [bend left=20] (t4a);
\path[arrow1] (t4a) edge [bend left=25] (t4);
\node[human] (t5) [right =of t4] {$\tau_5$};
\node[allocate] (t5a) [above right= 1.0cm and 1.25cm of t4] {$\tau_5^a$};
\path[arrow1] (t4) edge (t5);
\path[arrow1] (t4) edge [bend left=20] (t5a);
\path[arrow1] (t5a) edge [bend left=25] (t5);

\node[human] (t7) [right =of t6] {$\tau_7$};
\node[allocate] (t7a) [above right= 1.0cm and 1.25cm of t6] {$\tau_7^a$};
\path[arrow2] (t6) edge (t7);
\path[arrow2] (t6) edge [bend left=20] (t7a);
\path[arrow1] (t7a) edge [bend left=25] (t7);

\node[robot] (t8) [right =of t7] {$\tau_8$};
\path[arrow1] (t7) edge (t8);

\node[human] (t9) [right =of t8] {$\tau_9$};
\node[allocate] (t9a) [above right= 1.0cm and 1.25cm of t8] {$\tau_9^a$};
\path[arrow1] (t8) edge (t9);
\path[arrow1] (t8) edge [bend left=20] (t9a);
\path[arrow1] (t9a) edge [bend left=25] (t9);

\node[human] (t10) [right =of t9] {$\tau_{10}$};
\node[allocate] (t10a) [above right= 1.0cm and 1.25cm of t9] {$\tau_{10}^a$};
\path[arrow1] (t9) edge (t10);
\path[arrow1] (t9) edge [bend left=20] (t10a);
\path[arrow1] (t10a) edge [bend left=25] (t10);

\node[robot] (t12) [right =of t11] {$\tau_{12}$};
\path[arrow1] (t11) edge (t12);
\node[human] (t13) [right =of t12] {$\tau_{13}$};
\node[allocate] (t13a) [above right= 1.0cm and 1.25cm of t12] {$\tau_{13}^a$};
\path[arrow1] (t12) edge (t13);
\path[arrow1] (t12) edge [bend left=20] (t13a);
\path[arrow1] (t13a) edge [bend left=25] (t13);

\node[robot] (t14) [right =of t13] {$\tau_{14}$};
\path[arrow1] (t13) edge (t14);
\node[robot] (t15) [right =of t14] {$\tau_{15}$};
\path[arrow1] (t14) edge (t15);

\node[robot] (t17) [right =of t16] {$\tau_{17}$};
\node[error] (t17e) [above right= 1.0cm and 1.25cm of t16] {$\tau_{17}^e$};
\path[arrow2] (t16) edge (t17);
\path[arrow2] (t16) edge [bend left=20] (t17e);
\path[arrow1] (t17e) edge [bend left=25] (t17);
\node[human] (t18) [right =of t17] {$\tau_{18}$};
\node[allocate] (t18a) [above right= 1.0cm and 1.25cm of t17] {$\tau_{18}^a$};
\path[arrow1] (t17) edge [bend left=20] (t18a);
\path[arrow1] (t18a) edge [bend left=25] (t18);
\path[arrow1] (t17) edge (t18);
\node[robot] (t19) [right =of t18] {$\tau_{19}$};
\path[arrow1] (t18) edge (t19);
\node[human] (t20) [right =of t19] {$\tau_{20}$};
\path[arrow1] (t19) edge (t20);

\node[allocate] (t20a) [above right= 1.0cm and 1.25cm of t19] {$\tau_{20}^a$};
\path[arrow1] (t19) edge [bend left=20] (t20a);
\path[arrow1] (t20a) edge [bend left=25] (t20);

\node[state] (t21) [fill=black!60, thin, dashed, text=white, below right = 3cm and 1.5cm of t5] {$\tau_{21}$};
\path[arrow2] (t5) edge [bend left =20]  (t21);
\path[arrow2] (t10) edge [bend left =20]  (t21);
\path[arrow2](t15) edge [bend right =20]  (t21);
\path[arrow2] (t20) edge [bend right =20]  (t21);
\end{tikzpicture}
}
\caption{Temporary task graph of the experiment after task allocation.\\
Blue: Robot's tasks, Orange: Human's tasks, Cyan: Assigning tasks to the human, Red: Correcting human errors, Green: Already assigned tasks, Gray: Finished tasks}
\label{fig:allocatedtaskgraph}
\end{figure}

\textbf{Task scheduling}: After obtaining the optimal task allocation, the robot needs to find the optimal schedule by solving the optimization problem in \ref{eq:opt3}. 

After allocating tasks to both the robot, denoted as $\tau_{robot}$, and the human, denoted as $\tau_{human}$, such that $\tau_{new} = \tau_{robot} \cup \tau_{human}$, the robot proceeds to solve the optimization problem \ref{lin2} in order to find an optimal task schedule. Within this problem, the decision variables $S=\{s_{\tau_i}\}$ dictate the start times of the subtasks. Binary decision variables $O=\{o_{i,j}\}$ are employed to determine if subtask $\tau_i$ comes before or after $\tau_j$. The set $V$ comprises indices for all feasible subtasks in $\tau_{robot}$ that the robot can immediately perform by placing the blocks on the shared area. Moreover, binary decision variables $B=\{b_{\tau_i}\}$ are used to indicate whether $\tau_i \in V$ begins at $s_{\tau_i}=0$.

\begin{equation}
\min_{\left\{S, O, B\right\}} z \label{lin2}
\end{equation}   
 \vspace{-2em}
 \begin{align}
     &\text{subject to}\nonumber\\
    &\quad P\left(\tau_{i},\tau_{j}\right).f_{\tau_{i}}\leq s_{\tau_{j}}, &&\quad \forall \tau_{i},\tau_{j}\in \tau_{new}\\
    &\quad s_{\tau_i} - f_{\tau_j} + M(1-o_{i,j}) \geq 0 
    &&\quad \forall \tau_{i},\tau_{j}\in \tau_{human}\\
    &\quad s_{\tau_j} - f_{\tau_i} + Mo_{i,j} \geq 0 
    &&\quad \forall \tau_{i},\tau_{j}\in \tau_{human}\\
    &\quad s_{\tau_i} - f_{\tau_j} + M(1-o_{i,j}) \geq 0 
    &&\quad \forall \tau_{i},\tau_{j}\in \tau_{robot}\\
    &\quad s_{\tau_j} - f_{\tau_i} + Mo_{i,j} \geq 0 
    &&\quad \forall \tau_{i},\tau_{j}\in \tau_{robot}\\
    &\quad s_{\tau_i} - M b_{\tau_i} \leq 0
    &&\quad \forall i \in V\\
    &\quad \sum_{i\in V}{b_{\tau_i}} \leq \left|V\right|-1 \label{eq:opt5-const7}\\
    &\quad z - f_{\tau_i} \geq 0 
    &&\quad \forall \tau_{i} \in \tau_{new}
 \end{align}
 where M is a large positive constant. In this problem, we assume that rejecting the human agent's assignments and allocating tasks to the human takes zero seconds.

\textbf{Solving optimization problems:}  Our experiments were conducted on a computer running Ubuntu 18.04, equipped with an Intel Core i7-11700 CPU with 8 cores operating at 2.5GHz and 16 GB of RAM. After reformulating the task allocation and task scheduling problems, including their constraints, into mixed-integer linear programs, we utilized the GUROBI mathematical optimization solver, imposing a time limit to terminate the solver if it continues searching for additional solutions. We adopted a warm-start approach, providing the solver with a partially valid initial solution derived from the previous step's solution. Consequently, for the initial step, we could solve the problem offline.

\textbf{Updating $\alpha_f$ and $\alpha_e$}: 
In updating the robot's estimate of the human agent's preference to follow and their performance, the robot needs to detect changes and adapt its planning accordingly. That is, the estimation method has to be sensitive enough to consider these changes. However, it also must not be oversensitive to affect the planning abruptly. For example, a single mistake in the human's decision making will not be reacted upon precipitously. Taking advantage of the ``bounded memory adaptation model" approach, proposed in \cite{nikolaidis2017human}, we use a history of $k$-step in the past to estimate the human agent's preference and performance. A small value of $k$ makes the estimation sensitive to changes, and a large value leads to measuring the overall leading/following preference and performance. In this work, we chose $k=3$.

To update the belief, we employ the belief update method employed by \cite{nikolaidis2017human} to estimate human adaptability. According to this method, the system needs to be considered as a factorization of observable ($X$) and unobservable ($Y$) state variables of the system $S:X \times Y$. Subsequently, belief update can be computed as:
\begin{align}
    b'(y') = &\eta Z\big(x', y', a^R, o\big)\sum_{y\in Y}
    T_x\Big(x, y, a^R, a^H, x'\big) \label{eq:bupdate} \\ 
    &T_y \big(x, y, a^R, a^H, x', y'\big) 
    \pi^H\big(x, y, a^H\big) b\big(y\big), \nonumber 
\end{align}
where $T_x$ and $T_y$ are the transition functions, $z$ is the observation function, and $\pi^H$ is the human action model (policy).

\textbf{Following preference:} To update $\alpha_f$, we use Eq.~\ref{eq:bupdate}, where $y=\alpha_f$ and $b$ represents belief on $\alpha_f$. We consider $Z = 1$ and $T_x = 1$. We also consider $T_y \big(x, y, a^R, a^H, x', y'\big)=\mathbb{I}(y=y')$, where $\mathbb{I}$ is an indicator function. This is based on the reasonable assumption that the human agent's preference changes infrequently and is usually fixed. 
The human agent's strategy is determined based on the analysis of the preceding three steps in their actions. The actions taken into consideration by the robot for updating its policy regarding human following preferences include the actions of assigning a subtask to the robot ($F_1$), performing a subtask assigned by the robot ($F_2$), or refraining from performing a subtask assigned by the robot ($F_3$).
By denoting the occurrences of $F_1$, $F_2$, and $F_3$ as $f_1$, $f_2$, and $f_3$, respectively, within the sequence of human actions spanning a history of three steps, the resulting human policy can be defined as follows:

\begin{align}
    \pi^H_f(x,y, a^H) = \begin{cases}
    \frac{\alpha f_1 + f_2}{\alpha f_1 + f_2 + f_3}y &   a^H \in F_1 \cup F_2\\
    \frac{f_3}{\alpha f_1 + f_2 + f_3}y & a^H \in F_3
    \end{cases},
\end{align}
where $\alpha>1$ is a parameter that weighs cases where the human assigns a subtask to the robot more heavily. Therefore, based on Eq.~\ref{eq:bupdate}, the update of belief on $\alpha_f$ can be written as:
\begin{align}
    b^{\prime}(y^{\prime}) = \eta \sum_{y\in Y}
    \mathbb{I}(y^{\prime}=y) 
    \pi^H_f(y, a^H) b\big(y\big), 
\end{align}

\textbf{Human error:} To update $\alpha_e$, we utilize Eq.~\ref{eq:bupdate}, where $y=\alpha_e$, and $b$ denotes the belief on $\alpha_e$. We consider $Z = 1$ and $T_x = 1$. 

\begin{figure}[t]
      \centering
      \subfloat[Wrong action\label{fig:fol_lead1}]{%
        \includegraphics[scale=0.35]{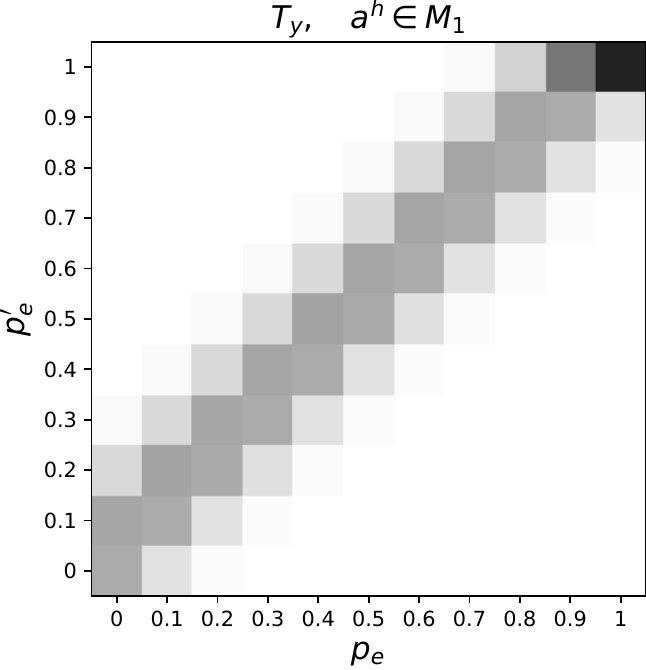}%
      }
      \hfill
      \subfloat[Correct action\label{fig:fol_lead2}]{%
        \includegraphics[scale=0.35]{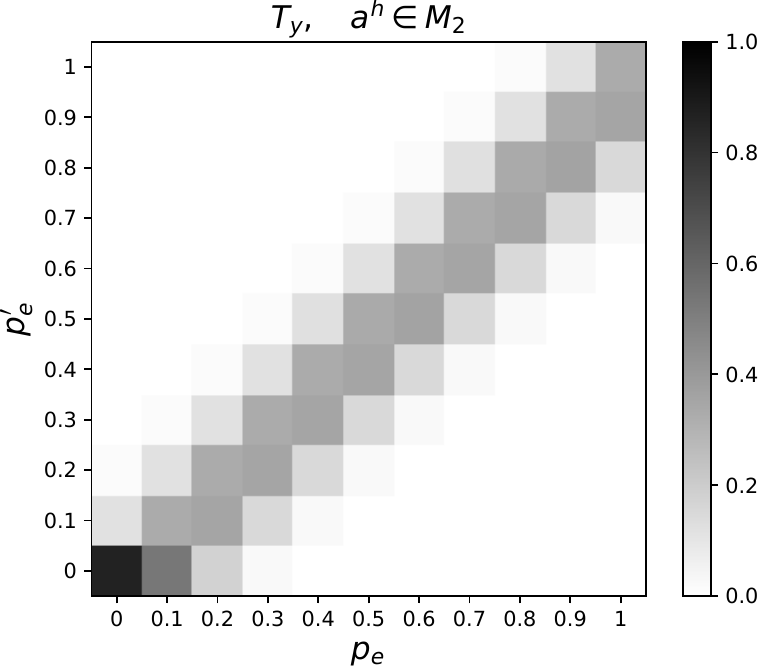}%
      }
      \caption{Estimating the human's error-proneness: Transition probability, $T_y$}
      \label{fig:heat}
    \end{figure}

Modeling human error, specifically the human memory model in this particular scenario, presents significant challenges and falls outside the primary focus of this paper. Nevertheless, we adopt a simplified model for $T_y$ and $\pi^H$ as necessary components for estimating $p_e$. We define two functions, $g_l(y)$ and $g_u(y)$, which respectively identify the closest value less than and greater than $y$ within the set $Y$. Consequently, we formulate $T_y$ as follows:
\begin{align}
 &T_y = 
 \begin{cases}
 \begin{aligned}
 p\big(y'\le &Z <g_u(y')\big) , \\ &Z \sim \mathcal{SN}\big(g_u(y),\,\sigma^{2}, \beta_1\big) 
 \end{aligned}    &\quad \text{ if } a^H \in M_1\\
 \begin{aligned}
 p\big(g_l(y'&)\le Z <y'\big) , \\ &Z \sim \mathcal{SN}\big(g_l(y),\,\sigma^{2}, \beta_2\big) 
 \end{aligned}   &\quad \text{ if } a^H \in M_2 
 \end{cases},
\end{align}
where $\beta$ is the skewness factor of the skew-normal distribution function $\mathcal{SN}(g_l(y),\,\sigma^{2}, \beta)$. The erroneous actions ($M_1$) and correct ones ($M_2$), if they are not assigned to the human by the robot, are taken into account for updating $\alpha_p$.  The transition probability $T_y$ heatmap is illustrated in Fig.~\ref{fig:heat}. Considering  $m_1$ and $m_2$ as the frequency counts of $M_1$, $M_2$, in the preceding three steps ($k=3$), the human error model is as follows:
\begin{align}
    \pi^H_e(x,y, a^H) = \begin{cases}
    \frac{m_2}{ m_1 + m_2}y &   a^H \in M_2\\
    \frac{m_1}{m_1 + m_2}y & a^H \in M_1
    \end{cases}.
\end{align}

Therefore, based on Eq.~\ref{eq:bupdate}, the update of belief on $\alpha_e$ can be written as:
\begin{align}
    b^{\prime}(y^{\prime}) = \eta \sum_{y\in Y}
    T_y \big(y, a^H, y'\big) 
    \pi^H_e(y, a^H) b\big(y\big). 
\end{align}

\subsection{Recruitment}
After obtaining ethics approval from the University of Waterloo Human Research Ethics Board, we initiated the participant recruitment process for the study by distributing recruitment flyers. This study involved three phases for each individual. 
\subsection{Study Procedure}
\subsubsection{Phase 0}
In this phase, after participants responded to our flyer, we emailed them a consent form and a questionnaire about their familiarity with and prior experience in robots and artificial intelligence.

\subsubsection{Phase 1 (In-Person)}
For this phase, we scheduled a 90-minute in-person session with each participant. In what follows, we explain the procedure.
\begin{enumerate}[label=\textbf{Step \arabic*:}, wide]
  \item After greeting participants, we explained the setup and showed them their work area. We used some slides to explain all the details and inform them that:
  \begin{itemize}
      \item the robot may make mistakes in its decision-making (deception),
      \item the team would be penalized \$1 for each misplaced block at the end of the task, once participants confirm task completion (deception).
  \end{itemize}
  \item (\textit{Task 0}) We asked them to complete the task alone, without the robot, for Pattern A. Before starting the task, they answered a question about their self-confidence to accomplish the task, and after doing the task, they completed a questionnaire about the task load (NASA-TLX, \cite{hart1988development}). Results from this step are not covered in this manuscript and are discussed in \cite{fetch_human}. 



  \item We asked them to watch a video\footnote{\url{https://youtu.be/ahZDo0_iyjg}} of the Fetch robot performing pick-and-place, and then answer the questionnaire about their trust in the robot (Muir’s questionnaire \cite{muir1996trust}).

  \item Participants worked with the GUI and practiced how to use it.
  \item (\textit{Task 1}) Based on the mode assigned to participants (permutation of patterns B, C, D), experimenters gave them the associated pattern (as a sheet of paper) and asked them to memorize it within 45 seconds. Then, they returned the pattern (e.g., $\text{B}_1$) and were given the second pattern, a partially known version of the first pattern (e.g., $\text{B}_2$). Next, prior to starting the task, we asked them to answer two questions regarding their self-confidence and the expected helpfulness of the robot. Afterwards, they started the task and the collaboration with Fetch. Having completed the task, they answered three sets of questionnaires regarding their perceived task load, trust, and perception of the robot.
  \begin{remark}
      Participants start the task first, and the robot waits for them. They can allocate subtasks to the robot. The robot starts working as soon as participants allocate a task to themselves. This allows the robot to initially update its belief about their following preference.
  \end{remark}
  \item (\textit{Task 2}) It followed the same procedure as \textit{Task 1}
  \item (\textit{Task 3}) It followed the same procedure as \textit{Task 1}

  \item  Finally, participants were asked to complete two sets of questionnaires. The first set focused on their performance as a team with the robot. The second set explored their collaborative experience using the short version of the User Experience Questionnaire (UEQ)\cite{ueqshort}. Additionally, participants were asked to rank the difficulty of tasks (Tasks 0-3) and respond to an open-ended question: \textit{``Which abilities would you improve or add to Fetch if you were to use it in a manufacturing setting?"}.
  
\end{enumerate}
The details and the results of the questionnaires on topics of participants' perception of the robot and collaboration (trust, helpfulness, task load, robot traits, team fluency, and user experience)  go beyond the scope of this manuscript and are reported in \cite{fetch_human}. 

\subsubsection{Phase 2 (Online)}

For the online phase, we prepared a video of each participant's collaboration with the robot, only for Pattern B. This video contained the synchronized videos showing the room from two different angles, a screen recording of the GUI, and Patterns $\text{B}_1$ and $\text{B}_2$. The video created for one of the participants is available online\footnote{\url{https://youtu.be/X6Rj0zwQhz8}}. During the online interview, we played the video and asked participants to discuss it, including their strategy, plan, and preferences during their collaboration in the task shown, as well as in two other tasks (Patterns C and D). Then, we asked them to complete two questionnaires about their leadership and followership styles. However, the results and analysis of these two questionnaires are not in the scope of this manuscript and are reported in \cite{fetch_human}. Finally, as per the approved ethics application, we explained about the \textit{``Deception"} elements in the study and asked them to sign another consent form to let us use their data. Participants were remunerated a \$30 gift card as an appreciation of their participation.

\section{Results \& Discussion}
We recruited 58 participants. However, we had to exclude data from 10 participants for various reasons, including a bug in the robot's program and the robot's failure. Consequently, our data analysis is based on the remaining 48 participants, consisting of 22 females, 24 males, and 2 selection ``others", with an average age of 24.02~$\pm$~3.93. The majority were university students (44), 3 were postdoctoral or visiting researchers, and 1 was a staff member. The results of this study can be analyzed from three key perspectives:  1- participants' perception of the tasks, the robot, and collaboration, 2- participants' preference and performance, 3- the robot's actions and performance. The first two perspectives were explored in \cite{fetch_human}. Building on their findings, this paper focuses on the latter. Furthermore, we delve into specific participant cases to illustrate how the robot adapted to individuals and various situations. Part of the data from this study is available online\footnote{\url{https://github.com/aslali/lead-follow-cobot-data}}.

\begin{remark}
We used the Kruskal–Wallis H test, a nonparametric statistical test, to determine whether there are statistically significant differences between two or more groups. When a significant overall difference exists among multiple groups, we employ the Dunn test as a post hoc analysis to identify specific group differences.
\end{remark}
\begin{remark}
    Analyzing the results based on the tasks corresponds to the chronological sequence, commencing from Task 0 and concluding with Task 3.
\end{remark}

\subsection{Highlights from Subjective \& Objective Analysis }
Here, we summarize the findings from analyzing 1- participants' perception of the tasks, the robot, and collaboration and 2- participants' preference and performance. These results are explored and elaborated in \cite{fetch_human}.
\begin{enumerate}
    \item Subjective assessments reveal an improvement in participants' perception of the robot and collaboration following their interaction, along with a reduction in perceived workload.
    \item Both subjective and objective assessments demonstrate that the robot effectively assisted participants in enhancing their performance and reducing errors.
    \item The interview results show that most of the participants preferred to take on the leading role and have more control over the robot. Based on participants' preferences, we categorized them into four groups (from highest leading preference to highest following preference): 1- lead (17 participants), 2- collaborative-lead (20 participants), 3- collaborative-follow (4 participants), and 4- follow (3 participants).
    \item According to the interviews, participants found the robot to be slower than themselves and preferred to handle more blocks. 

    \item The results indicate that participants, in general, found Pattern B more difficult compared to Patterns C and D. In addition, there was the highest number of participants who made at least one mistake in Pattern B. This can be attributed to the fact that Pattern B was both challenging to memorize and had the most unknown spots, leading participants to rely on the robot.
\end{enumerate}

\subsection{Robot's Actions and Estimation}
We briefly highlighted participants' preferences and performance based on the interviews and recorded data from the user study. However, we must also explore how the robot could adapt to participants' preferences and performance.

\subsubsection{Participants Preference} 
The robot updated its estimation of the human preference based on the $3$-step history of the human's actions. 
The robot needs to consider the human agent's preference changes and adapt accordingly. However, we needed to create a measure to evaluate the robot's performance in estimating the human overall preference. To do so, first, we normalize the completion time and then fit a polynomial (e.g., degree 4) on the estimated values, $f(t), t\in\left[0, 1 \right]$. Then, we calculate the area under the curve in a certain range, as a measure of participants' overall estimated preference,
\begin{equation}
    oep = \int_{t_0}^1 f(t),
\end{equation}
where we considered $t_0 = 0.2$.

\begin{figure}
    \centering
    \includegraphics[width=0.9\linewidth]{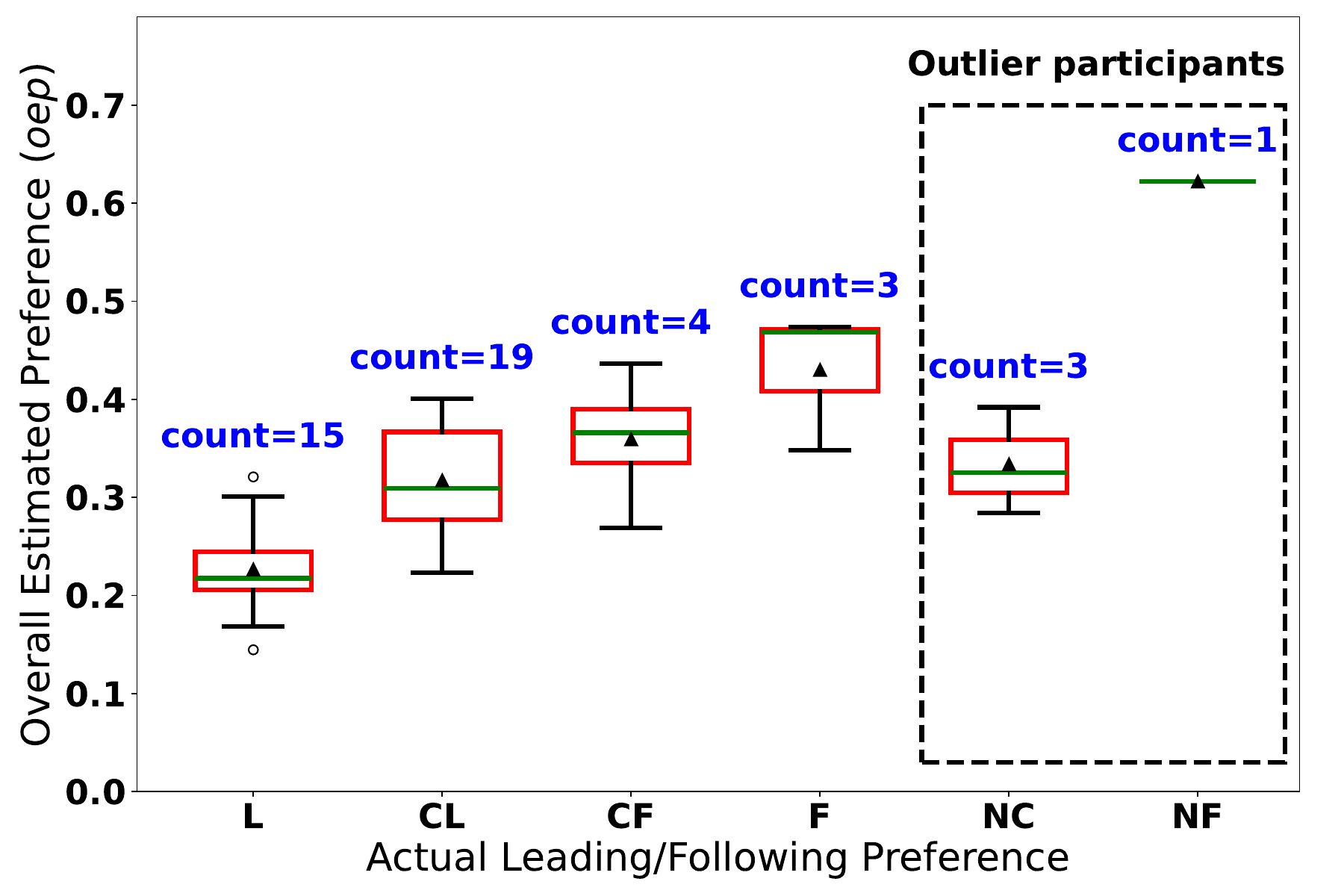}
    \caption{The robot performance in estimating participants' preference by comparing with the participants' actual preference, gathered through interviews. The numbers above each bar show the number of participants falling into that group. Lead (\textbf{L}),  Collaborative-Lead (\textbf{CL}),  Collaborative-Follow (\textbf{CF}), and  Follow (\textbf{F}), Neither-Collaborative (\textbf{NC}), Neither-Follow (\textbf{NF})}
    \label{fig:estimate_real_pref}
\end{figure}

To evaluate the robot's ability to estimate participants' preference, for each of them, we measured the average of the overall estimated preference ($oep$) in Tasks 1, 2, and 3. Combining them with the information gathered in interviews (i.e., participants' actual preferences) leads to Fig~\ref{fig:estimate_real_pref}, which shows that the robot effectively estimated participants' actual preferences. Four participants did not fit into any of the four identified groups. One participant, classified as neither-follow, preferred working independently, choosing not to lead or follow but to complete tasks on his own. He did not assign any tasks to Fetch and completed all the tasks assigned by the robot. The other three participants, referred to as neither-collaborative, began their tasks quickly to avoid forgetting the pattern. They primarily selected tasks for themselves but were also open to completing those assigned by Fetch. These two groups are highlighted within a dashed rectangle in  Fig~\ref{fig:estimate_real_pref}.

 In addition, we analyzed participants'  overall estimated preference for each task and noticed no significant difference among them ($H(2) = 2.29$, $p=0.32$). This is justifiable as the majority of participants preferred to lead the robot.

 \subsubsection{Task Difficulty - Following Preference}

\begin{figure}
\centering
\subfloat[\label{fig:preference_pattern}]{%
    \includegraphics[width=0.47\linewidth]{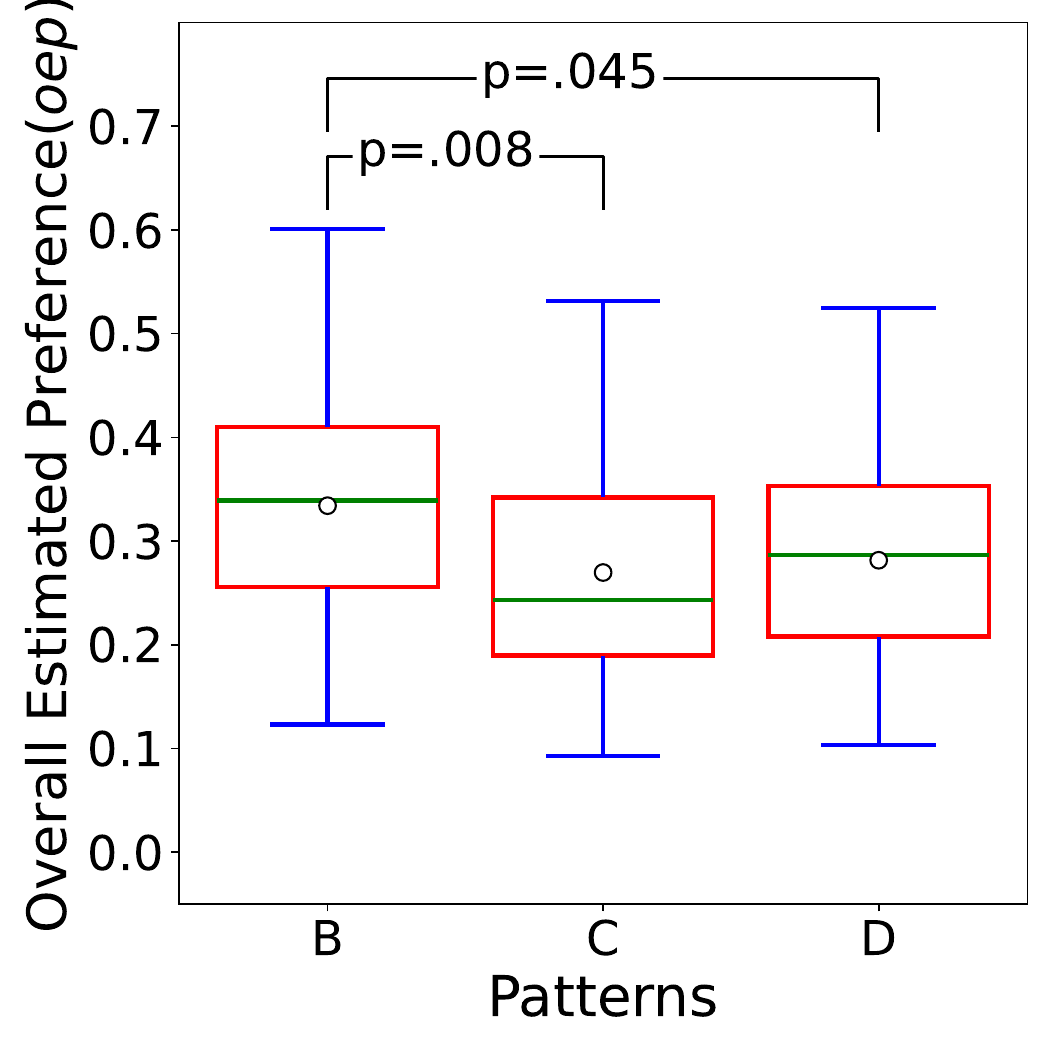}}
    \subfloat[\label{fig:tasks_assigned_robot}]{%
    \includegraphics[width=0.47\linewidth]{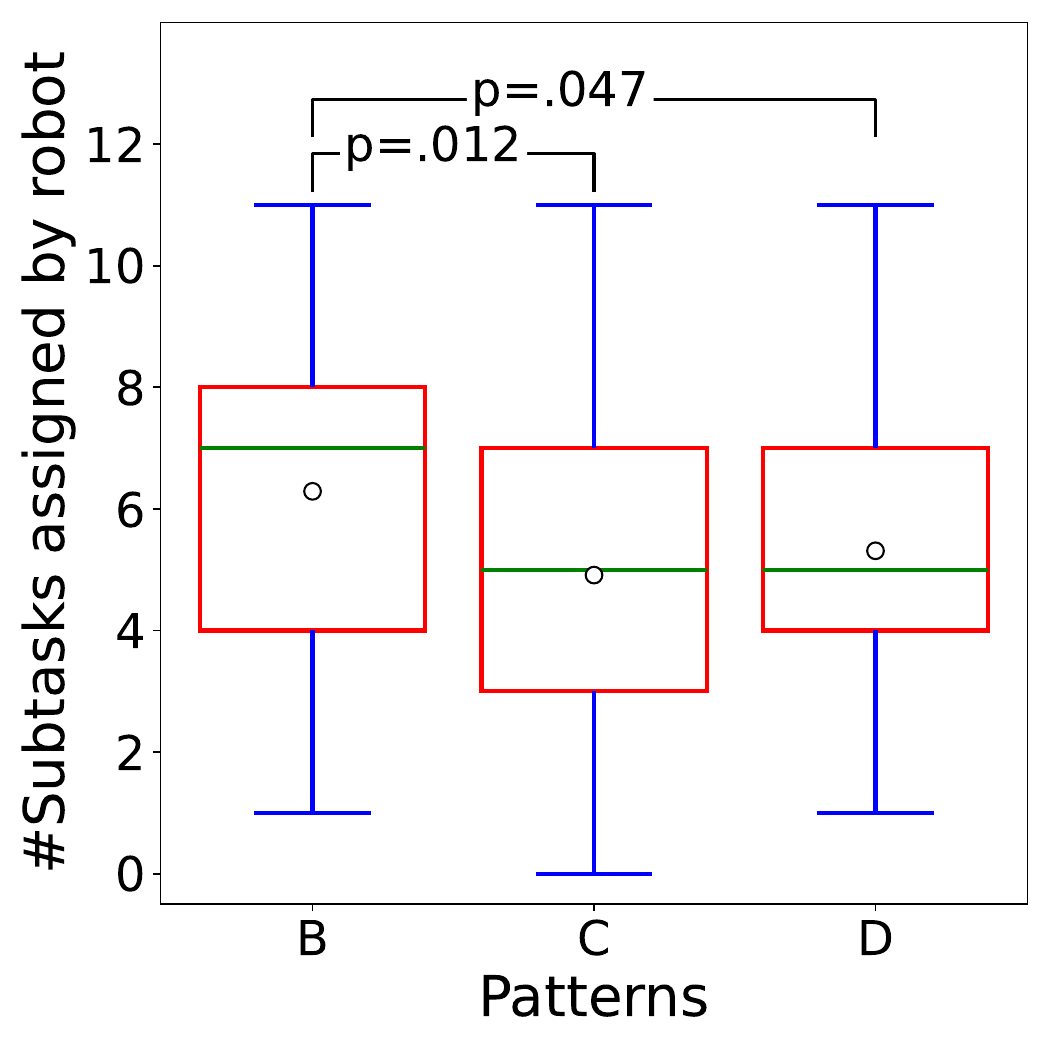}}
  \caption{\textbf{a)} Participants' overall estimated leading/following preference based on the patterns. The robot's estimation of participants' preference to follow was significantly higher in Pattern B compared to Patterns C and D. \\
  \textbf{b)} Subasks assigned to participants by the robot based on the patterns. The number of assigned subtasks by the robot to participants was significantly higher in Pattern B compared to Patterns C and D. }
\end{figure}

 As stated in the highlights of the participants' subjective and objective measurements, participants found Pattern B more challenging than Patterns C and D. Analyzing participants'  overall estimated preference based on the patterns, as illustrated in Fig.~\ref{fig:preference_pattern}, shows that there is a significant difference in participants' following preference based on the patterns ($H(2) = 6.5$, $p=0.039$), indicating a significantly higher preference for following in Pattern B compared to Patterns C and D. 


Based on the robot's task planning algorithm, the greater the human agent's preference for following or the occurrence of errors, the more tasks are assigned to them by the robot. Fig.~\ref{fig:tasks_assigned_robot} shows that the robot significantly assigned more subtasks to the human in Pattern B compared to Patterns C and D ($H(2) = 6.32$, $p=0.042$). Likewise, this is justifiable as Pattern B was more difficult than the two others, and participants made more mistakes or preferred to follow the robot, and the robot could accordingly adapt its planning. 

\subsubsection{Task Assignment \& Distribution}

\begin{figure*}
    \centering
    \includegraphics[width=0.7\linewidth]{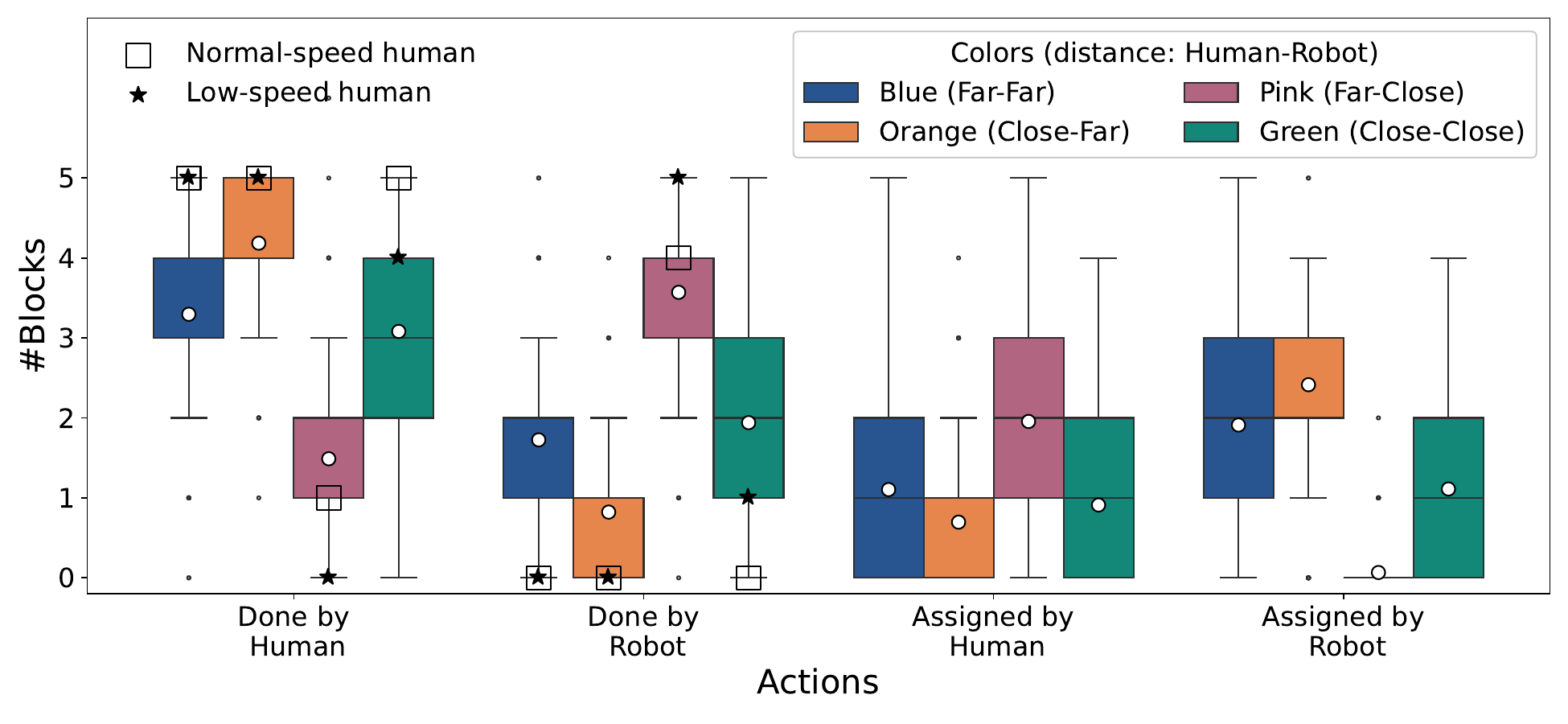}
    \caption{Number of subtasks completed and assigned by each agent for each block color. The optimal offline task allocations, based solely on completion time, are shown for two different human speeds.}
    \label{fig:blocks_dist_exp}
\end{figure*}

Referring back to the distribution of the blocks (Table~\ref{tab:color_dist}) and the lower speed of the robot compared to the human, the optimal task allocation will be closer to assigning pink blocks to the robot and blue and orange blocks to the human. Fig.~\ref{fig:blocks_dist_exp} shows the colors of blocks completed by participants and those assigned by the robot. This distribution, resulting from the interplay between the robot and participants, is close to optimal. To determine the offline optimal task allocation, focusing solely on minimizing time and disregarding human preference and performance, we considered two possible human speeds: a normal speed of approximately $1.2 m/s$ and a slower speed of $0.7 m/s$. In the normal speed scenario, the optimal solution for all three patterns is to assign only four pink subtasks to the robot, with the remaining tasks given to the human. In the slow speed scenario, all five pink subtasks and one green subtask are assigned to the robot, while the rest are handled by the human. As expected, most of the orange blocks were completed by the human agent due to participants' expected rational decision-making and the robot's assignments. Similarly, the robot completed pink subtasks. Regarding the blue subtasks, on average, participants completed most of them. However, the distribution of assigned and completed blue subtasks by the robot and human ranges from 0 to 5. This variation is due to some participants, as indicated by interviews, who preferred to assign more blue tasks to the robot, reducing their physical effort at the expense of longer collaboration time.

\begin{figure}
    \centering
    \includegraphics[width=1\linewidth, height=0.55\linewidth]{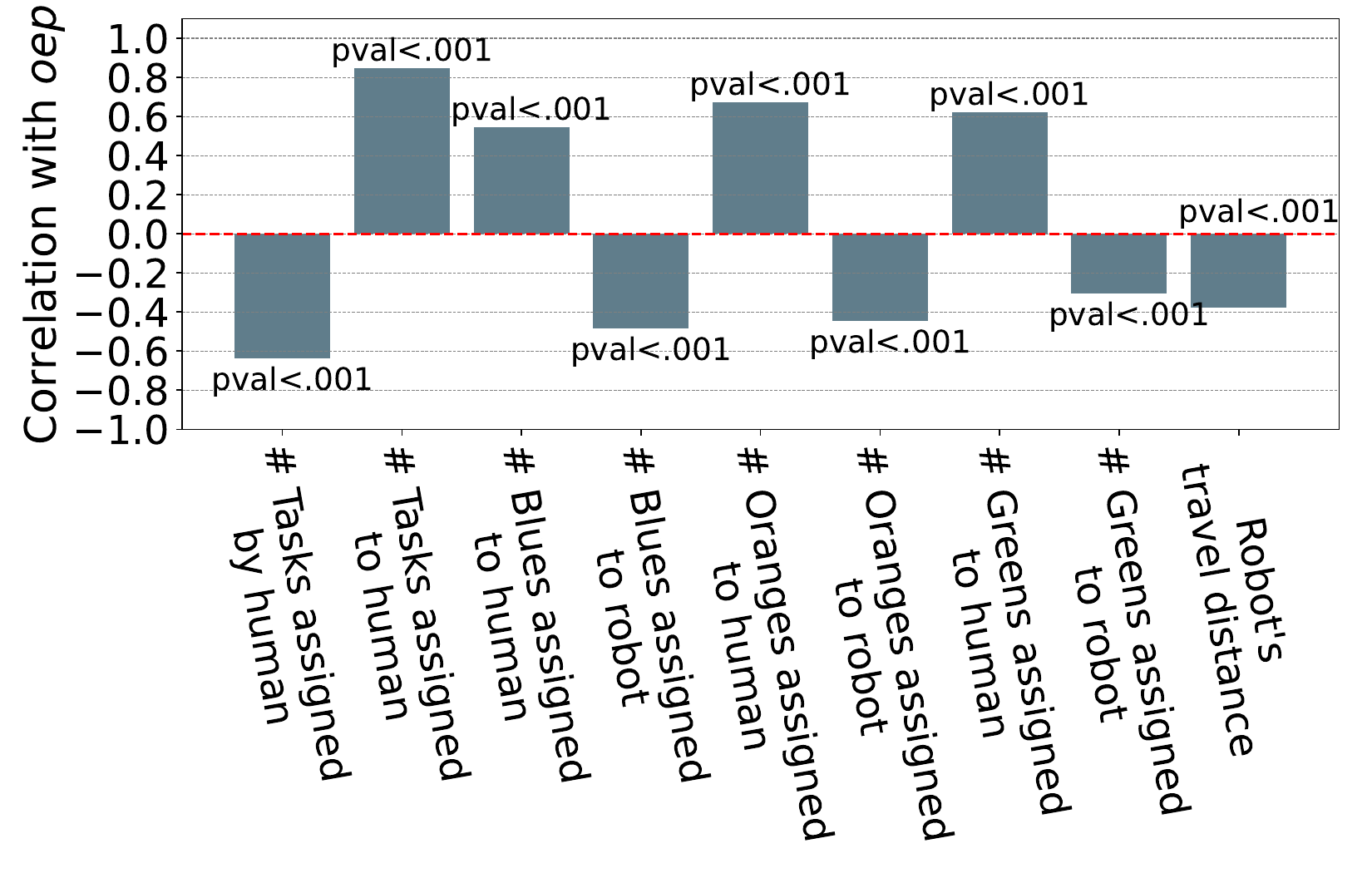}
    \caption{Correlation between participants' overall estimated preferences and the number of subtasks assigned by both the robot and participants.}
    \label{fig:assign_est_follow}
\end{figure}

\textbf{Following preference vs. Task distribution:} 
Based on the designed algorithm, we expect the robot to assign more subtasks to a human with the following preference, which is particularly important for blue subtasks. Additionally, we are interested in examining how the robot's estimation of participants' following preference relates to the number of subtasks participants assign to it. As shown in Fig.~\ref{fig:assign_est_follow}, there is respectively a strong positive correlation and a strong negative correlation between the number of subtasks the robot assigned to participants and the number of subtasks that the human assigned to the robot with the estimated participants' following preference. This aligns with the designed algorithm.

Furthermore, we can observe a moderate negative correlation between the number of assigned blue and orange tasks to the robot by humans and their following preference. Conversely, a moderate to strong positive correlation exists between the orange, green, and blue blocks allocated to the human agent and the estimated following preference. A weak negative correlation exists between the number of assigned green tasks and participants' following preference. As expected, the correlation between the approximate robot's travel distance and the estimated following preference shows a moderate negative correlation, as when the robot has a more leading role, it assigns tasks that are farther from itself to the human agent. 

\textbf{Discussion:} 
The results indicate that the robot could successfully identify participants' following preferences and adapt its planning accordingly. We also observed that the robot's estimation of participants' following preference was higher for Pattern B than for Patterns C and D, resulting in more tasks being assigned to participants. This aligns with the previous finding that Pattern B was more challenging than the other two patterns, and participants required more support from the robot. All of these, in concert, showed that the robot was more of a leader in challenging tasks. This could result from either the participants' choice to follow the robot or the significant number of mistakes they made. Additionally, the robot could guide participants toward optimal task allocation to minimize collaboration time, taking into account the block's location and the fact that it is slower in comparison to the human agent.

\subsection{Participant-Specific Analysis}
We reviewed participants' overall preferences and performance, but another goal of the framework is to track changes in these over time. To assess its online adaptation and planning ability, we focus on specific participants' results.

\begin{figure*} 
    \centering
    \subfloat[Case 1\label{fig:result_cases1_1}]{%
    \includegraphics[width=0.31\linewidth]{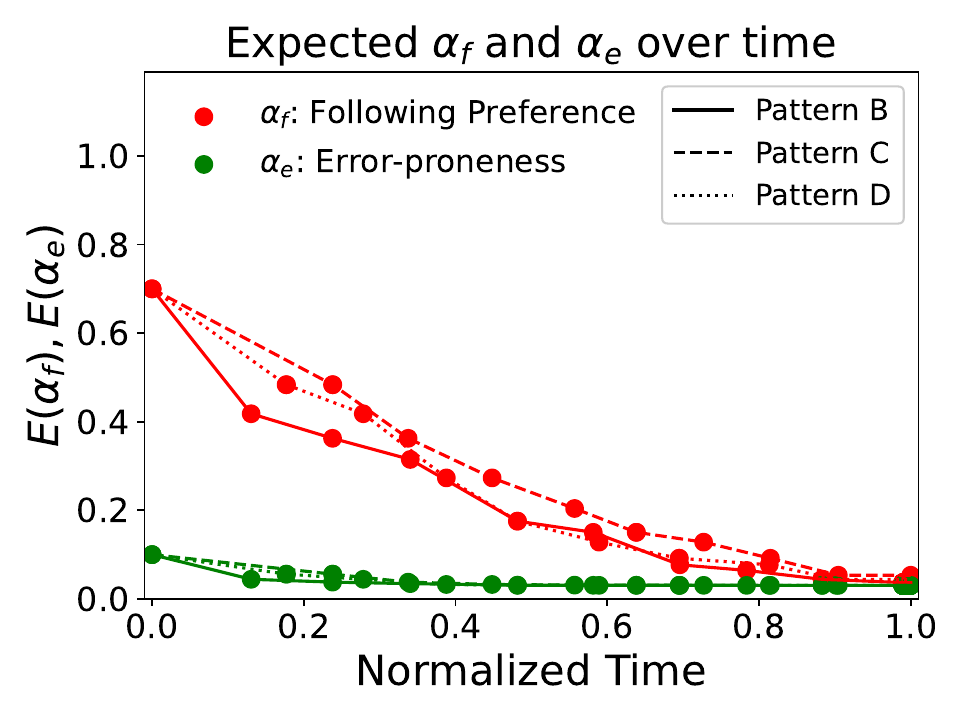}}
    \subfloat[Case 2\label{fig:result_cases1_2}]{%
    \includegraphics[width=0.31\linewidth]
    {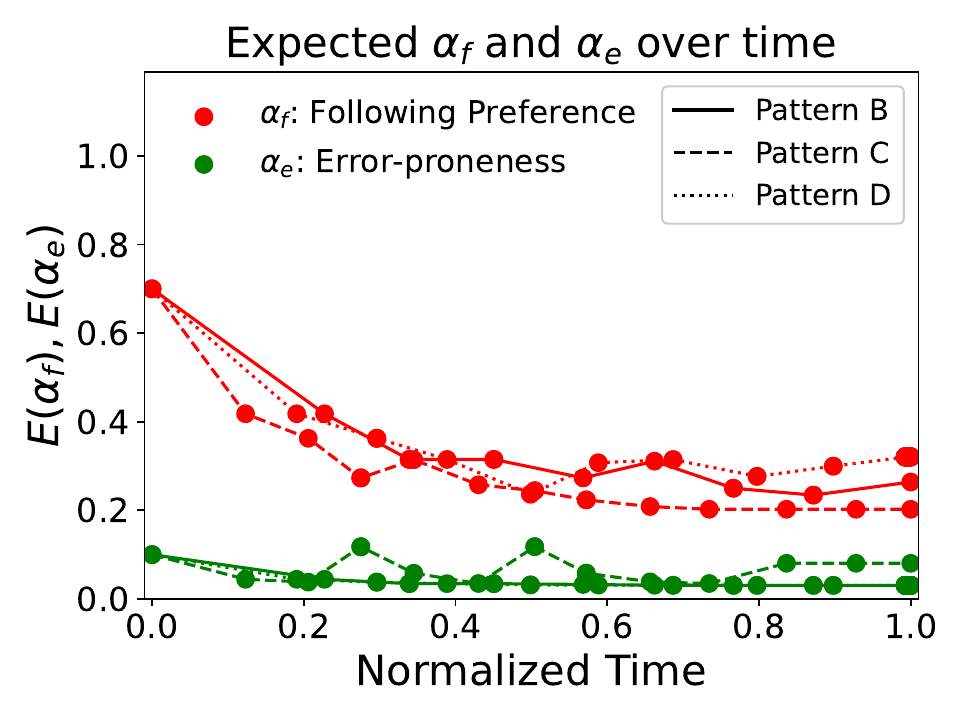}}
    \subfloat[Case 3\label{fig:result_cases2_1}]{%
    \includegraphics[width=0.31\linewidth]{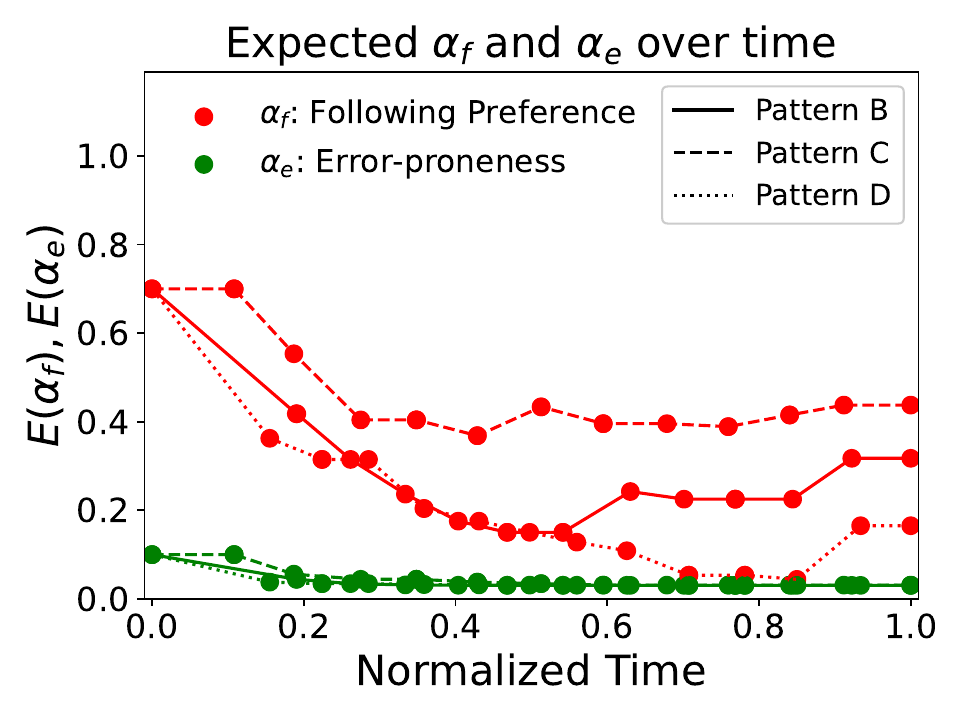}}
    \hfill
    \subfloat[Case 4\label{fig:result_cases2_2}]{%
    \includegraphics[width=0.31\linewidth]{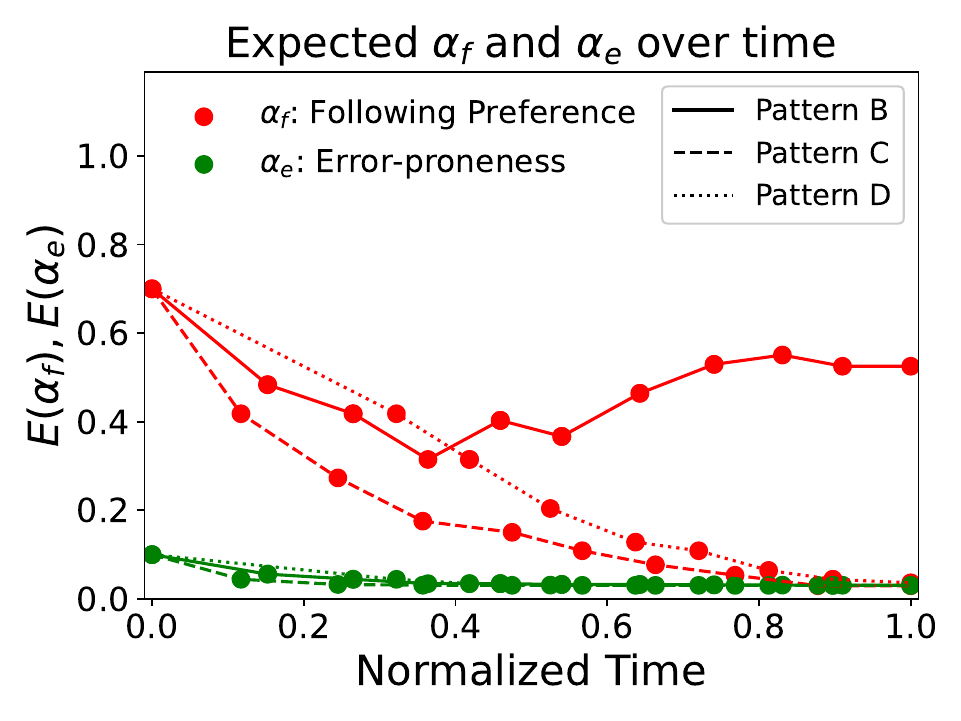}}
    \subfloat[Case 5\label{fig:result_cases4_1}]{%
    \includegraphics[width=0.31\linewidth]{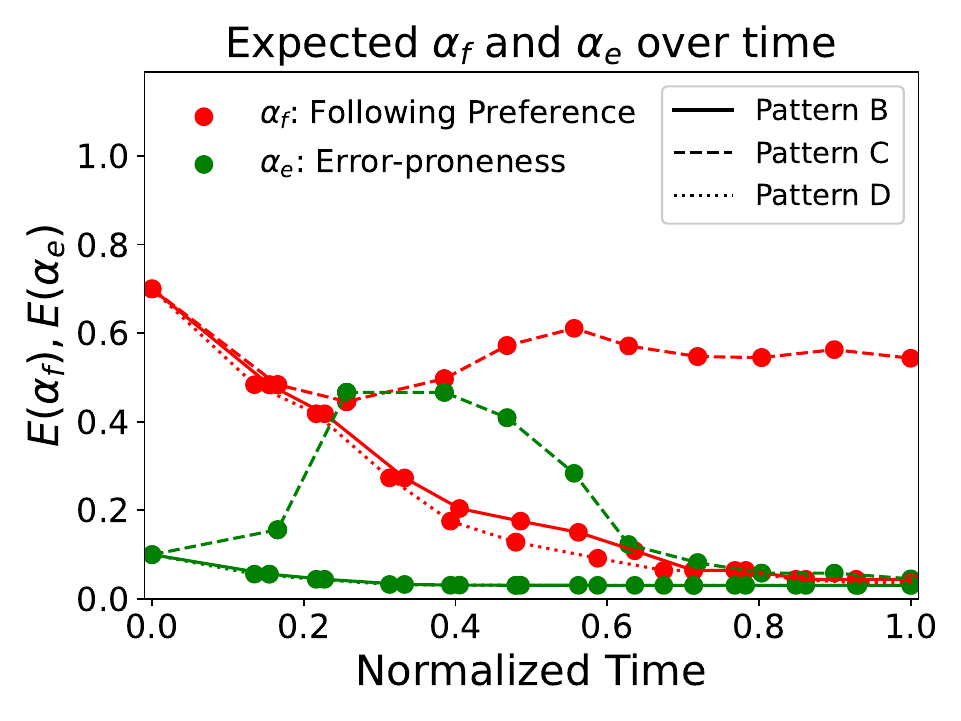}}
    \subfloat[Case 6\label{fig:result_cases4_2}]{%
    \includegraphics[width=0.31\linewidth]{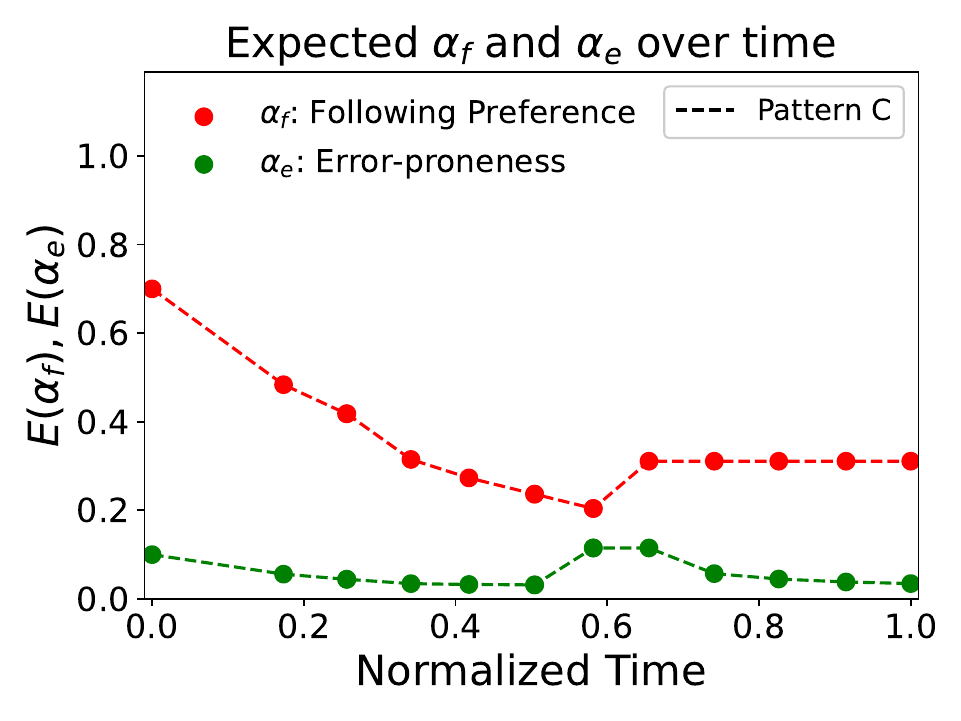}}
    \hfill
    \subfloat[Case 7\label{fig:result_case3}]{%
    \includegraphics[width=0.31\linewidth]{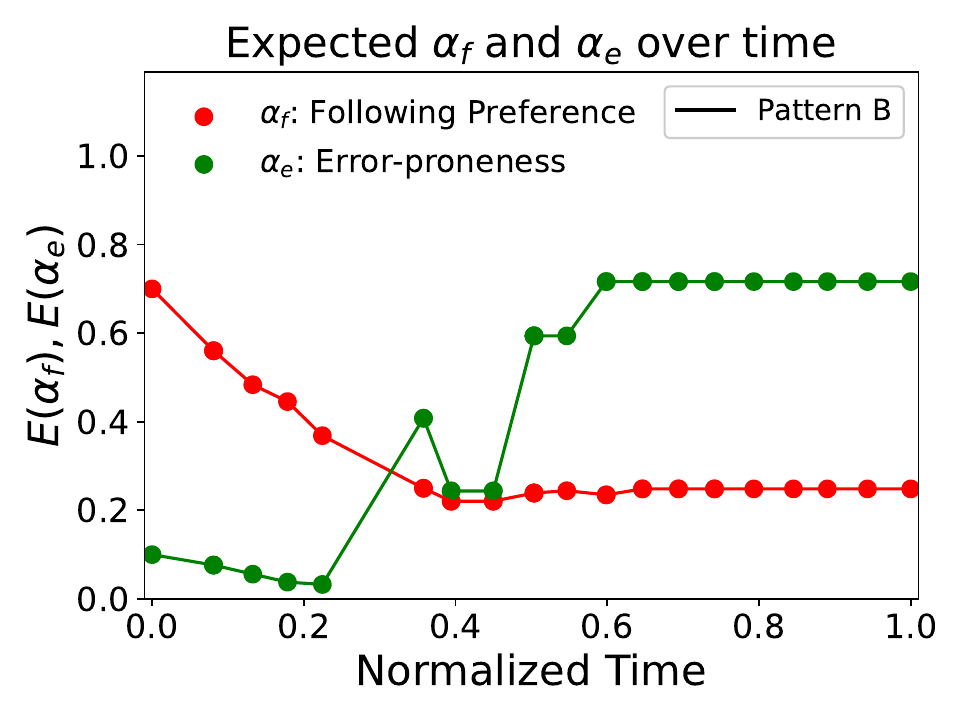}}
    \subfloat[Case 8\label{fig:result_cases5_1}]{%
    \includegraphics[width=0.31\linewidth]{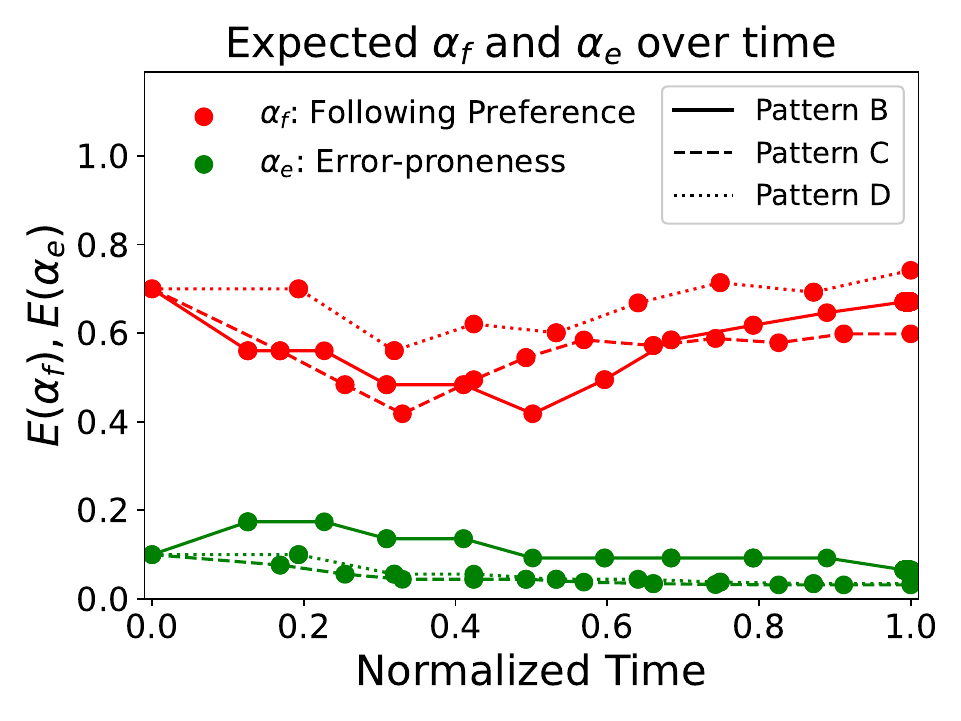}}
    \subfloat[Case 9\label{fig:result_cases5_2}]{%
    \includegraphics[width=0.31\linewidth]{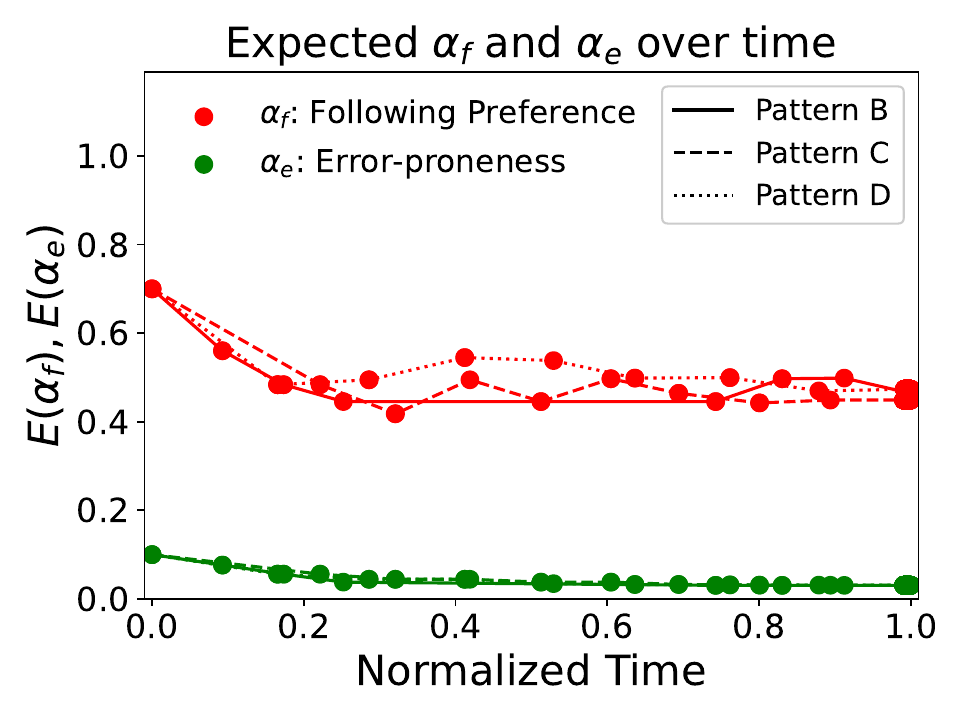}}
  \caption{The robot's estimates of the participants' preference and performance: \textbf{a, b)}  leading or collaborating-leading preference and high accuracy (Cases 1 and 2); \textbf {c, d)} leading preference with occasional robot support (Cases 3 and 4); \textbf {e, f)} leading preference with occasional errors (Cases 5 and 6); \textbf {h)} leading preference with sudden performance drop (Case 7); \textbf {i, j)} following or collaborating-following preference (Cases 8 and 9)}
  \label{fig:case1_partic} 
\end{figure*}

\emph{Cases (1\&2) Leading or collaborating-leading preference with a high accuracy: } 
These participants preferred to lead the team and had a high accuracy. Thus, the robot estimated their leading preference and gave them the leading role. Fig.~\ref{fig:result_cases1_1} and \ref{fig:result_cases1_2} show the robot estimate of two participants' error-proneness and following preference, with respectively leading {(\textit{case 1})} and collaborating-leading preferences {(\textit{case 2})}.

\emph{{Cases (3\&4)} Leading preference with occasional robot support:}
{One participant (\textit{case 3}) preferred to lead while minimizing physical effort by assigning most tasks to the robot. In their interview, they confirmed using this  strategy across all tasks. However, the participant forgot part of Pattern B and let the robot assign them some tasks. They were also uncertain about the last row of Pattern C. The robot’s estimates of their preference and performance are shown in Fig.\ref{fig:result_cases2_1}. Fig.\ref{fig:result_cases2_2} shows estimates for another participant (\textit{case 4}) with a leading preference who was uncertain about Pattern B.
}

\emph{{Cases (5\&6)} Leading preference with  occasional errors: }
{These two participants took the leading role and generally recalled the pattern, but made occasional mistakes. The robot detected these errors and increased its estimate of their error-proneness ($\alpha_e$). Figures~\ref{fig:result_cases4_1} and~\ref{fig:result_cases4_2} illustrate the robot's estimation of their preference and performance. In Fig.~\ref{fig:result_cases4_1},  the participant (\textit{case 5})  made several consecutive mistakes initially but improved their performance after robot assistance.}

\emph{{(Case 7)} Leading preference with sudden performance drop:} {This participant led the robot in Pattern B and performed well until near the end of the task. They became confused about the final row and made mistakes, insisting on incorrect decisions despite the robot rejecting their assignments and correcting them. The robot then updated its belief about the participant’s performance (Fig.~\ref{fig:result_case3}) and, while still considering them to have a leading preference, reassumed the lead role and guided them to the correct pattern by assigning more tasks.}

\emph{{Cases (8\&9)} Following or collaborating-following preference: } 
Fig.~\ref{fig:result_cases5_1} and \ref{fig:result_cases5_2} show the robot estimate of two participants' error-proneness and following preference, with following {(\textit{case 8})} or collaborating-following preference {(\textit{case 9})}. Therefore, the robot took the leading role and assigned subtasks to them.

\section{Simulation Study }
{In our user study, key human behaviors often occurred infrequently, making it difficult to comprehensively evaluate the algorithm’s performance across varied scenarios, especially with a limited number of participants and time-intensive setups. Despite the valuable insights of real-world human-robot collaboration, its limitations necessitate a simulation study.}
For brevity, the details of the study and collaborative scenario are omitted,  they can be found in \cite{noormohammadi2022task}, and the simulation codes are available online\footnote{\url{https://github.com/aslali/lead-follow-cobot-simulation}}. In this paper, we focus on the following aspects and present some new analysis:
\begin{enumerate*}
    \item The robot's planning algorithm's sensitivity to different levels of human preference and performance.
    \item {Comparison of the adaptive and non-adaptive methods.}
\end{enumerate*}

\subsection{Sensitivity Analysis}
\begin{figure*}
    \centering
    \subfloat[$P_{follow}=0.9$\label{fig:simul_pfollow9}]{%
    \includegraphics[width=0.32\linewidth]{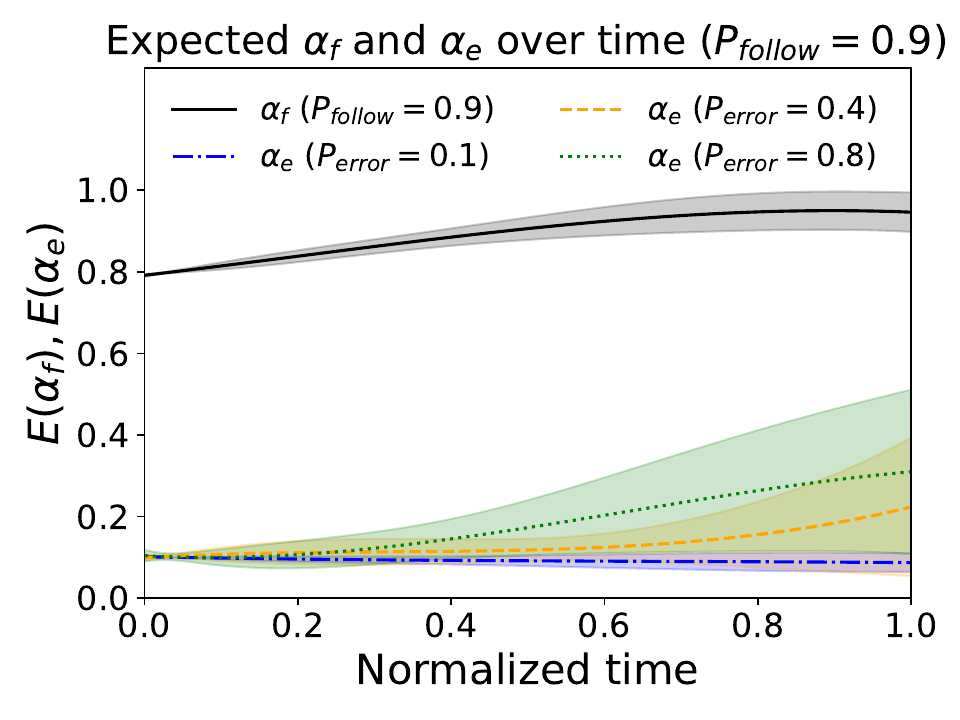}
    }
    \subfloat[$P_{follow}=0.6$\label{fig:simul_pfollow6}]{%
    \includegraphics[width=0.32\linewidth]{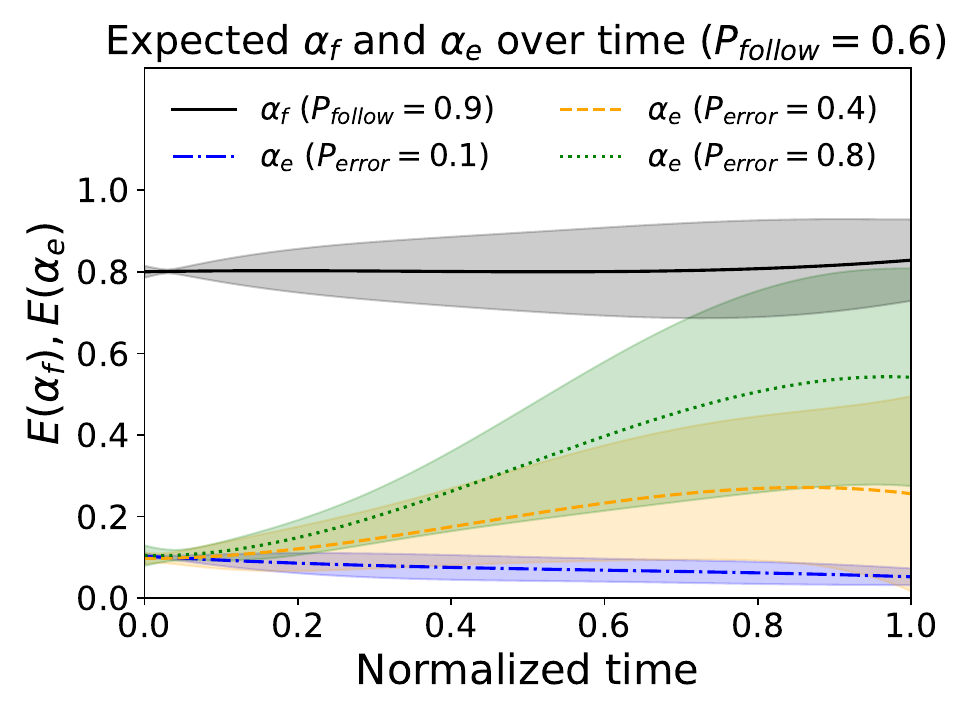}%
    }
    \subfloat[$P_{follow}=0.0.3$\label{fig:simul_pfollow3}]{%
    \includegraphics[width=0.32\linewidth]{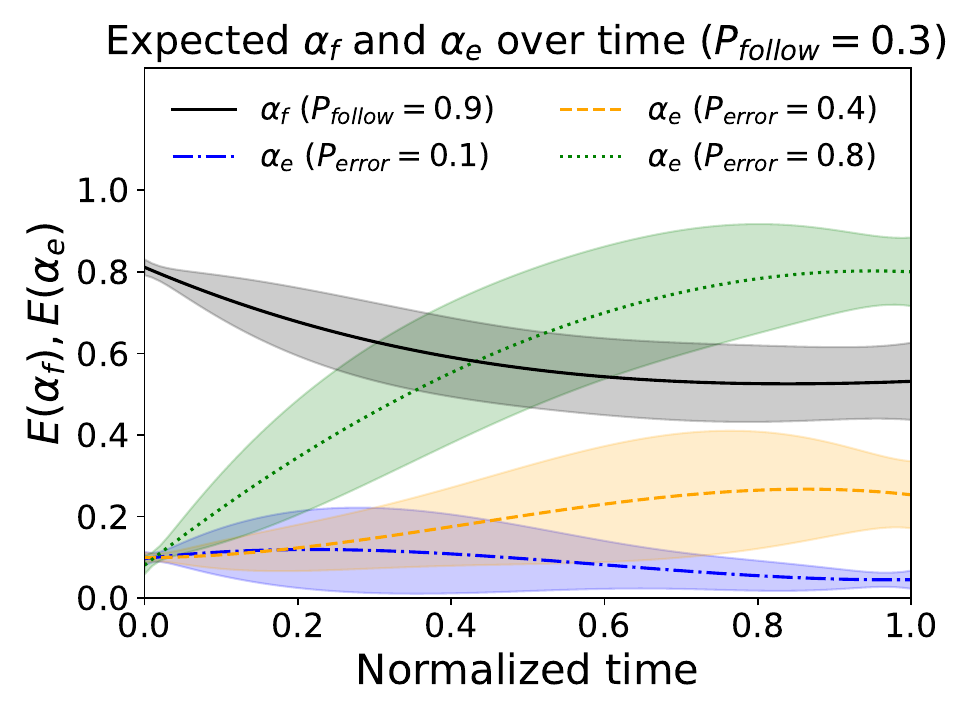}%
    }

      \caption{The robot's estimates of the human agent's following preference and error-proneness with \textbf{a)} strong following preference, \textbf{b)} moderate following preference, and \textbf{c)} slight following preference.}
      \label{fig:simul_pfollow}
\end{figure*}

In this simulation, we created a simplified human behavior using two key parameters: the preference to follow the robot $P_{\textit{follow}}$ and the error-proneness $P_{\textit{error}}$. Then, we explore nine different combinations of the human agent's following preference -- ranging from strong ($P_{\textit{follow}}=0.9$), moderate ($P_{\textit{follow}}=0.6$), and slight ($P_{\textit{follow}}=0.3$) -- alongside three different levels of error-proneness, low ($P_{\textit{error}}=0.1$), moderate ($P_{\textit{error}}=0.4$), and high ($P_{\textit{error}}=0.8$). Fig.~\ref{fig:simul_pfollow} displays the robot's estimates of the human agent's preferences and error-proneness, categorized by different values of $P_{follow}$.

\begin{itemize}[leftmargin=*]
\item{Strong preference to follow the robot ($\mathbf{P_\textit{follow}=0.9}$):} {Here, the robot takes the lead and directs the team's actions (Fig.~\ref{fig:simul_pfollow9}). As a result, even with potentially low human accuracy, mistakes remain minimal due to reliance on the robot's guidance.}

\item{Moderate preference to follow the robot ($\mathbf{P_\textit{follow}=0.6}$):} {In this case,  the robot recognizes this (Fig.~\ref{fig:simul_pfollow6}) and assigns fewer tasks to the human than in the ${P_\textit{follow}=0.9}$ case. However, as the human's error rate increases, the robot allocates more tasks to maintain team performance.}

\item{Slight preference to follow the robot ($\mathbf{P_\textit{follow}=0.3}$):} {When the human prefers to lead, the robot adapts by assigning fewer tasks (Fig.~\ref{fig:simul_pfollow3}). If the human’s accuracy drops, the robot increases task assignments to regain the lead and support team performance.}
\end{itemize}

\begin{figure*}
      \centering
      \subfloat[Low error-proneness\label{fig:ncml1}]{%
        \includegraphics[width=0.31\linewidth]{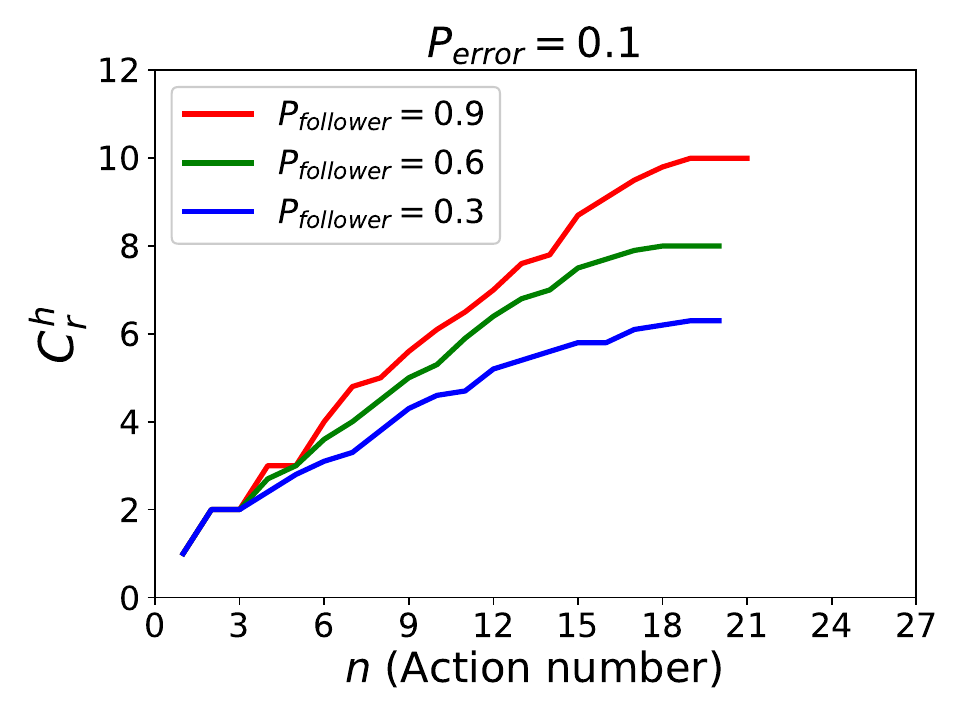}%
      }
      \hfill
      \subfloat[Moderate error-proneness\label{fig:ncml4}]{%
        \includegraphics[width=0.31\linewidth]{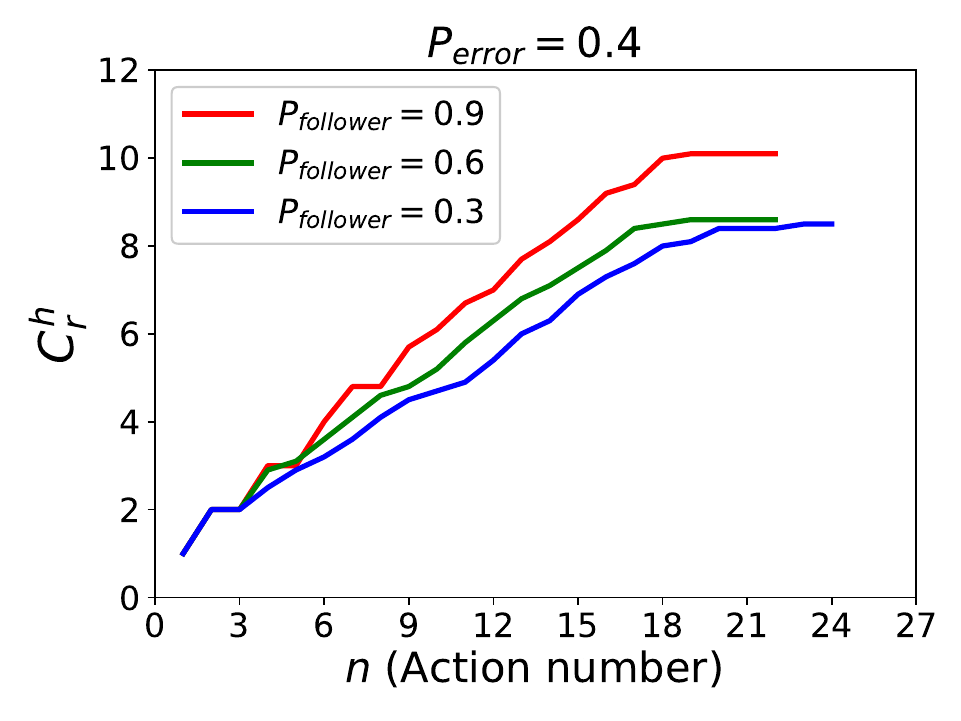}%
      }
    \hfill
      \subfloat[High error-proneness\label{fig:ncml8}]{%
        \includegraphics[width=0.31\linewidth]{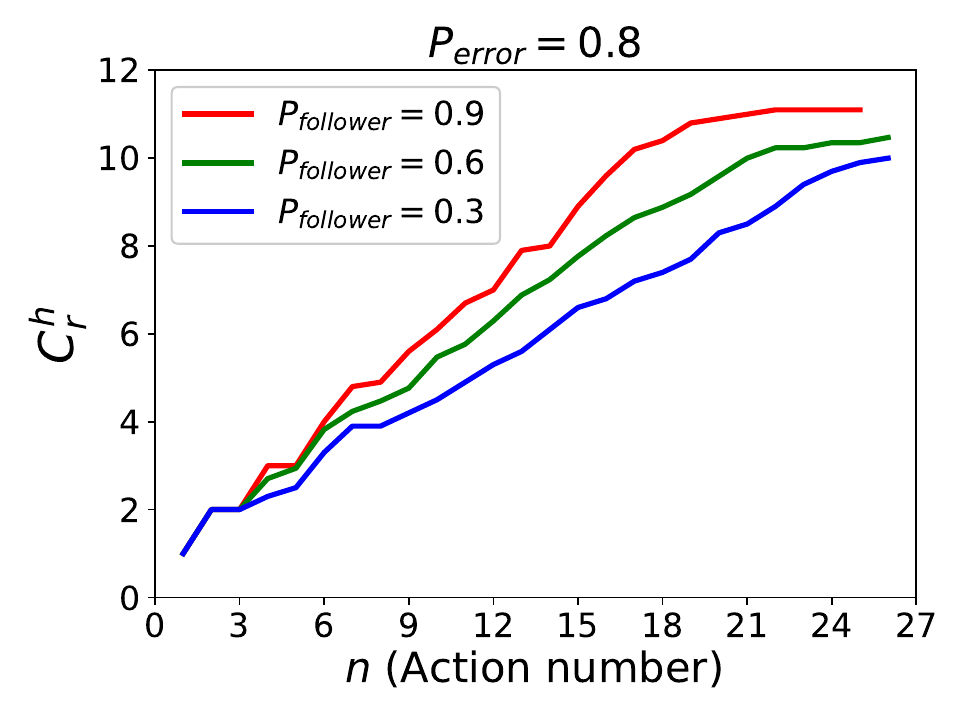}%
      }
      \caption{The cumulative number of subtasks assigned to the human by the robot ($C^h_r$), grouped by different levels of error-proneness.}
      \label{fig:ncml}
\end{figure*}

\emph{Assigned tasks by the robot:} 
{Fig.\ref{fig:ncml} shows the cumulative number of subtasks assigned to the human ($C_r^h$) for each error-proneness level. Figures\ref{fig:ncml1}–\ref{fig:ncml8} illustrate that the robot assigns more subtasks as the human’s following preference ($P_{follow}$) increases. Within the same following preference (same colors), higher error-proneness ($P_{error}$) leads to more assignments, especially when $P_{follow} = 0.3$.}

\subsection{Adaptive vs. Non-adaptive Approaches}

{We compare two approaches: the proposed adaptive method and a non-adaptive method, where the robot does not adjust its planning based on the human’s preferences or performance. The non-adaptive method assumes the human follows its lead ($\alpha_f = 0$) and focuses solely on minimizing task completion time. We consider two beliefs the robot may hold about human error-proneness: 1) the human makes no errors ($\alpha_e = 0$), and 2) the human has a fixed error-proneness of $\alpha_e = 0.35$.}

{We examine all nine combinations of human leading preference and error-proneness, as shown in Tables~\ref{tbl: simul_merror} and~\ref{tbl: simul_massign}, with each scenario simulated 15 times. Table~\ref{tbl: simul_merror} reports the average number of incorrect human actions, and Table~\ref{tbl: simul_massign} shows the average number of subtasks assigned to the human by the robot. In both tables, values for the non-adaptive and adaptive methods appear in the same cell, separated by a horizontal line—non-adaptive above, adaptive below. Statistically significant differences identified via ANOVA are highlighted.}

{$P_{\text{follow}} = 0.3$:
When $P_{\text{error}} = 0.1$ or $0.4$ and the robot ignores the human’s preference for leading, it tends to assign more subtasks to the human, which can lead to distrust or dissatisfaction in real-world settings. When the human is highly error-prone ($P_{\text{error}} = 0.9$) and prefers to lead, the adaptive algorithm assigns more subtasks to reduce the human’s agency and thereby minimize errors. This difference is less pronounced when the robot already assumes a fixed error-proneness of 0.35, as it accounts for potential mistakes regardless of adaptation. Conversely, if the robot assumes a fixed error-proneness of 0.35 but the human prefers to lead and performs well, it still assigns many subtasks.}

{$P_{\text{follow}} = 0.6$: The adaptive algorithm performs similarly to the non-adaptive one, as the non-adaptive approach assumes the human follows. Although not statistically significant, the results suggest the adaptive method detects lower human performance and assigns more subtasks accordingly.}


{$P_{\text{follow}} = 0.9$: There is no significant difference between the adaptive and non-adaptive approaches, as the latter assumes the human follows. Since the human follows, even with high error-proneness, few mistakes occur. The adaptive method results in a slightly higher number of errors, as the robot initially gives the human more agency while learning their preferences and performance, leading to a small, non-significant increase in incorrect actions.}

{Overall, the results show that relying on a fixed initial belief without adaptation leads to suboptimal robot behavior when it misaligns with the human’s actual preference or performance. The non-adaptive method may also fail if the human’s preference or performance changes during the task. These scenarios, however, are not addressed in this simulation and could be addressed in future work.}

      
\definecolor{lightgray}{gray}{0.9}
\begin{table*}
\begin{minipage}[t]{0.49\textwidth}
    \centering
    \caption{Mean number of incorrect human actions in a comparative simulation study: adaptive (bottom) vs. non-adaptive (top) planning.}

    \renewcommand{\arraystretch}{1.1} 
    \setlength{\tabcolsep}{5pt}  
    \begin{tabular}{|c|c|c|c|c|c|c|}
        \hline
        \multirow{2}{*}{$P_\textit{follow}$} & \multicolumn{3}{c|}{$P_{error}$ ($\alpha_e =0$)} & \multicolumn{3}{c|}{$P_{error}$ ($\alpha_e =0.35$)} \\
        \cline{2-7}
         &  \textbf{0.1} &  \textbf{0.4} &  \textbf{0.8} &  \textbf{0.1} &  \textbf{0.4} &  \textbf{0.8}\\
        \noalign{\global\arrayrulewidth=0.6pt} 
        \cline{2-7}
        \noalign{\global\arrayrulewidth=0.4pt}
        \hline
        \textbf{0.3}  
            & \multicolumn{1}{!{\vrule width 0.4pt}c|}
            {\begin{tabular}{c} 1.07 \\ \hline 1 \end{tabular}}  
            & \begin{tabular}{c} 4.53 \\ \hline 2.8\end{tabular}  
            & \cellcolor{lightgray} \begin{tabular}{c} 10.93 \\ \hline 9.2 \end{tabular}  
            & \begin{tabular}{c} 1.13 \\ \hline 1 \end{tabular}  
            & \begin{tabular}{c} 3.8 \\ \hline 2.8 \end{tabular}  
            & \begin{tabular}{c} 7.86 \\ \hline 9.2 \end{tabular}  \\
        \hline
        \textbf{0.6}  
            &\multicolumn{1}{!{\vrule width .4pt}c|}
            {\begin{tabular}{c} 0.27 \\ \hline 0.47 \end{tabular}}  
            & \begin{tabular}{c} 2.67 \\ \hline 1.87 \end{tabular}  
            & \begin{tabular}{c} 4.87 \\ \hline 4 \end{tabular}  
            & \begin{tabular}{c} 0.67 \\ \hline 0.47 \end{tabular}  
            & \begin{tabular}{c} 1.73 \\ \hline 1.87 \end{tabular}  
            & \begin{tabular}{c} 4.8 \\ \hline 4 \end{tabular}  \\
        \hline
        \textbf{0.9}  
            &\multicolumn{1}{!{\vrule width 0.4pt}c|}
            {\begin{tabular}{c} 0.2 \\ \hline 0.2 \end{tabular}}  
            & \begin{tabular}{c} 0.6 \\ \hline 0.78 \end{tabular}  
            & \begin{tabular}{c} 1.33 \\ \hline 1.67 \end{tabular}  
            & \begin{tabular}{c} 0 \\ \hline 0.2\end{tabular}  
            & \begin{tabular}{c} 0.2 \\ \hline 0.78 \end{tabular}  
            & \begin{tabular}{c} 1.27 \\ \hline 1.67 \end{tabular}  \\
        \hline
    \end{tabular}
    \label{tbl: simul_merror}
\end{minipage}
\hfill
\begin{minipage}[t]{0.49\textwidth}
    \centering
    \caption{Mean number of Subtasks Assigned to the Human by the robot in a comparative simulation study: adaptive (bottom) vs. non-adaptive (top) planning.}
    \renewcommand{\arraystretch}{1.1} 
    \setlength{\tabcolsep}{5pt}
    \begin{tabular}{|c|c|c|c|c|c|c|}
        \hline
        \multirow{2}{*}{$P_\textit{follow}$} & \multicolumn{3}{c|}{$P_{error}$ ($\alpha_e =0$)} & \multicolumn{3}{c|}{$P_{error}$ ($\alpha_e =0.35$)} \\
        \cline{2-7}
         &  \textbf{0.1} &  \textbf{0.4} &  \textbf{0.8} &  \textbf{0.1} &  \textbf{0.4} &  \textbf{0.8}\\
        \noalign{\global\arrayrulewidth=0.60pt} \cline{2-7}
        \noalign{\global\arrayrulewidth=0.4pt}
        \hline
        \textbf{0.3}  
            & \multicolumn{1}{!{\vrule width 0.4pt}c|}
            {\cellcolor{lightgray}\begin{tabular}{c}  8.93 \\ \hline 6.8 \end{tabular}}  
            & \cellcolor{lightgray}\begin{tabular}{c} 9.67 \\ \hline 8\end{tabular}  
            & \begin{tabular}{c} 11.07 \\ \hline 10.07 \end{tabular}  
            & \cellcolor{lightgray}\begin{tabular}{c} 7.8 \\ \hline 6.8 \end{tabular}  
            & \begin{tabular}{c} 8.87 \\ \hline 8 \end{tabular}  
            & \begin{tabular}{c} 9.5 \\ \hline 10.07 \end{tabular}  \\
        \hline
        \textbf{0.6}  
            &\multicolumn{1}{!{\vrule width .4pt}c|}
            {\begin{tabular}{c} 7.73 \\ \hline 8.47 \end{tabular}}  
            & \begin{tabular}{c} 8.47 \\ \hline 8.53 \end{tabular}  
            & \begin{tabular}{c} 9.87 \\ \hline 10.19 \end{tabular}  
            & \begin{tabular}{c} 8 \\ \hline 8.47 \end{tabular}  
            & \begin{tabular}{c} 9.86 \\ \hline 8.53 \end{tabular}  
            & \begin{tabular}{c} 10.07 \\ \hline 10.19 \end{tabular}  \\
        \hline
        \textbf{0.9}  
            &\multicolumn{1}{!{\vrule width 0.4pt}c|}
            {\begin{tabular}{c} 9.86 \\ \hline 9.86 \end{tabular}}  
            & \begin{tabular}{c} 9.6 \\ \hline 10.0 \end{tabular}  
            & \begin{tabular}{c} 10.13 \\ \hline 10.67 \end{tabular}  
            & \begin{tabular}{c} 9.93 \\ \hline 8.86\end{tabular}  
            & \begin{tabular}{c} 9.93 \\ \hline 10.0 \end{tabular}  
            & \begin{tabular}{c} 10.53 \\ \hline 10.67 \end{tabular}  \\
        \hline
    \end{tabular}
    \label{tbl: simul_massign}
\end{minipage}
\end{table*}

\section{Conclusion and future work}
We investigated how proactive task planning and allocation can improve human-robot collaboration efficiency. Previous work often overlooked either human agents' leading/following preferences or their performance. Our study focuses on balancing these factors while maintaining effective collaboration and a positive human perception of the robot.

Based on interviews with participants, we categorized them, based on their following/leading preference into four groups: ``lead", ``collaborative-lead", ``collaborative-follow", and ``follow", with the majority falling into the first two categories. This means that participants would prefer to take on more of a leading role and have more control over the collaboration. This finding can guide the design of collaborative scenarios and collaborative robots. Furthermore, we compared this result, showing actual participants' leading/following preferences, with the robot's estimation of their preference. This analysis showed that the robot successfully inferred their preference in most cases.

The results indicated that for more difficult tasks, participants trusted the robot more than their own abilities, which led them to take relatively more following roles. Our proposed task planning method properly inferred this need and provided more help to participants by taking on more leading roles. The robot could also identify when participants struggled to remember the correct patterns and made errors, and accordingly, it fixed their errors and provided more help. 

We analyzed the distribution of subtasks between the robot and human agent, showing that overall their interplay led to near-optimal task allocation. The results indicated a moderate to strong correlation between estimated participants' preferences and measures of task distribution. Additionally, participants were assigned more subtasks in difficult tasks as they preferred to follow the robot or made many errors, prompting the robot to reassume the lead.

In summary, we have developed a planning architecture for a robot, allowing it to adjust to its human teammate's preferences and performance while updating its plan in response to the task state and the human agent's actions. The findings indicate that most participants favored assuming a more leading role and exerting control over the team. The results also show the robot's capability to adapt its planning to provide assistance when required by its human teammate.


\subsection{Generalizability}

Although assessing the adaptive task planner required us to employ it in a specific scenario focused on the kitting task, this framework can be applied to various collaborative scenarios requiring task allocation and scheduling with appropriate customization. For instance, in a co-assembly task similar to those outlined in \cite{schmidbauer2023empirical, lee2022task, lamon2019capability}, our framework enables the robot to create efficient plans and allocate tasks based on its awareness of limitations while considering the performance (e.g., speed and accuracy) and preferences (such as whether humans prefer to assign tasks or be assigned tasks, or their preferred types of tasks) of human collaborators. As long as human preferences do not significantly hinder team efficiency, the robot can accommodate them. The framework is also designed to adapt to changing human preferences; for instance, a human might transition from assigning tasks to favoring robot-allocated tasks when faced with challenging tasks or experiencing fatigue.

Similar to the outcomes of our user study discussed in this paper and in \cite{fetch_human}, we can expect improvements in participants' satisfaction, trust, and overall perception of collaboration, along with reduced perceived task load when applying the adaptive planner to new scenarios. The planner’s ability to adapt to varying human performance and preferences, due to factors like fatigue or task difficulty, could further enhance collaboration. Incorporating error detection would also ensure robust performance, even when human errors occur. Based on this and some other studies in \cite{chanseau2019does, schmidbauer2023empirical, roy2019automation}, future designs should account for humans' desire to retain control, as this can improve the effectiveness of human-robot systems.

\subsection{Limitations and Future Work}
{This work has some limitations in study design and methods. Most participants were young adults from the University of Waterloo, whereas the intended users are working adults in environments like manufacturing and warehouses—groups that may have different expectations and perceptions of robot teammates. Recruiting participants from those settings could help design more practical collaborative scenarios and robots. Although we tried to simulate a working environment, e.g.,  by including  pick-and-place tasks, an autonomous mobile manipulator robot, a conveyor belt, and safety equipment, future work could further enhance realism by modeling settings such as warehouse automation or assembly. Another key aspect of manufacturing is the occurrence of sudden or unpredictable events that only humans can manage; incorporating such scenarios would bring the study closer to real-world conditions.}

{An important direction for future work is extending the framework to handle conflicts between agents, such as when both believe their decisions are correct. This study assumed the robot's decisions were always accurate. However, in real-world scenarios, the robot may make mistakes it is unaware of, due to issues like perception errors. Also, we used the human's correctness in selecting block colors as a measure of accuracy, which was easily understood by participants and used to assess performance. Some scenarios, however, may require alternative metrics, such as completion time or travel distance, which may be less comprehensible or measurable for humans. This mismatch in how team performance is interpreted by the robot and human may lead to conflicts.}

As another avenue for future work, to reduce the adaptive approach's dependence on accurate estimations of human preferences and performance, it can be enhanced with robust/stochastic methods and  recovery techniques to address potential misestimations. Additionally, the task planning framework presented here focuses on single-human, single-robot scenarios, while real-world scenarios often involve team structures like single-human, multi-robot or multi-human, multi-robot configurations. Future research could extend the framework to accommodate these multi-agent scenarios.


\bibliographystyle{IEEEtran}
\bibliography{ref}
\appendices \label{append}
\section*{APPENDIX}

\subsection*{Autonomous Pick-and-Place System Overview} \label{setup}
This section provides an overview of the robot stack used for autonomous pick-and-place tasks in the user study.

\subsubsection{Robot's Base Planning}

The Fetch robot, equipped with a LiDAR sensor, uses the slam\_karto package with pose graph SLAM for 2D mapping. For localization, we utilize the AMCL package’s Adaptive Monte Carlo method. Base motion planning combines Dijkstra’s algorithm for global paths and the trajectory rollout (TR) algorithm for local adjustments, both provided by the move\_base package. However, near the tables, the standard planner is bypassed, and direct velocity commands are issued for precise positioning.

\subsubsection{Block Detection}
We use ArUco markers with unique IDs to label each foam block. The Fetch robot's RGBD camera provides depth and color data for detecting the markers. When a marker is identified, the robot maps the block's location to the global map. To ensure accuracy, the robot continuously updates the block locations after each observation, considering both the distance at the time of the last detection and the elapsed time since. Errors in measurement increase with greater distance from the marker and with time due to potential block movement, so frequent updates help minimize these distance-based and time-related errors, improving accuracy. When picking up a specific block, the robot adjusts its position and orientation using the marker data to ensure accurate alignment for grasping.

\subsubsection{Pick-and-Place Task}

For successful pick-and-place tasks, the block must be within the robot's reachability. The robot continuously tracks the target block by adjusting its head to maintain a clear line of sight. Once the block is within reach, RRT$^*$ is used to plan the arm's motion, followed by a five-step pick-up process: 1- moving the arm above the block, 2- lowering it into the pick-up position, 3- grasping the block, 4- lifting it, and 5- retracting the arm toward the robot’s body for safe transport. The same sequence is used for placing the block. Since the Fetch robot lacks force sensors, joint effort (torque) is monitored to detect when the block touches the table during placement.

\begin{IEEEbiography}
[{\includegraphics[width=1in,height=1.25in,clip,keepaspectratio]{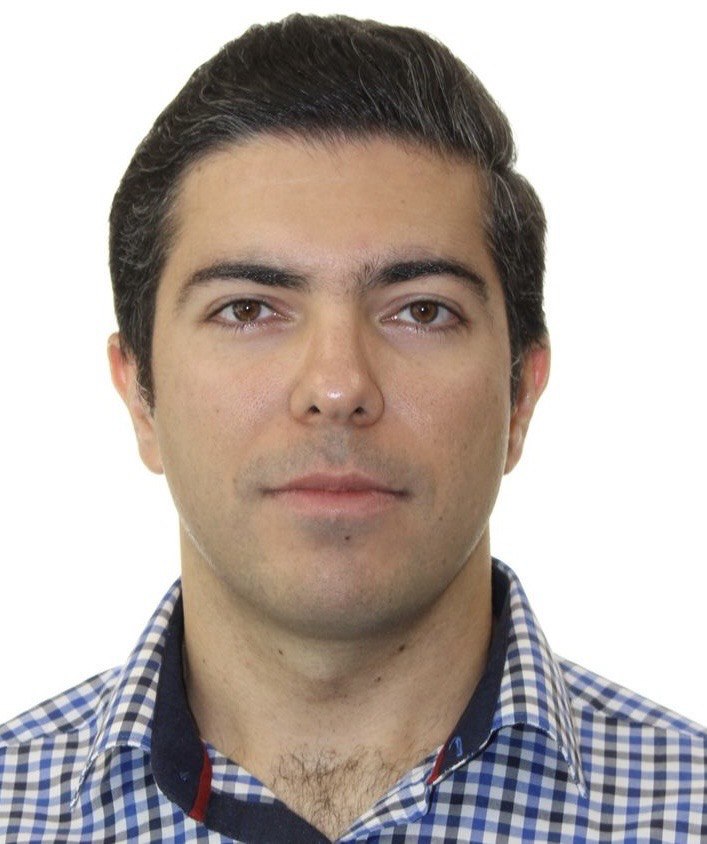}}]{Ali Noormohammadi Asl}
(Member, IEEE) received the bachelor’s and master’s degrees in electrical engineering, focusing on systems and controls from the Amirkabir University of Technology, Tehran, Iran, in 2012, and K. N. Toosi University of Technology, Tehran, Iran, in 2017, and the Ph.D. degree in systems design engineering from the University of Waterloo, Waterloo, Canada, in 2024. 

His research interests include motion planning, human-robot collaboration, and multirobot coordination.
\end{IEEEbiography}
\begin{IEEEbiography}[{\includegraphics[width=1in,height=1.25in,clip,keepaspectratio]{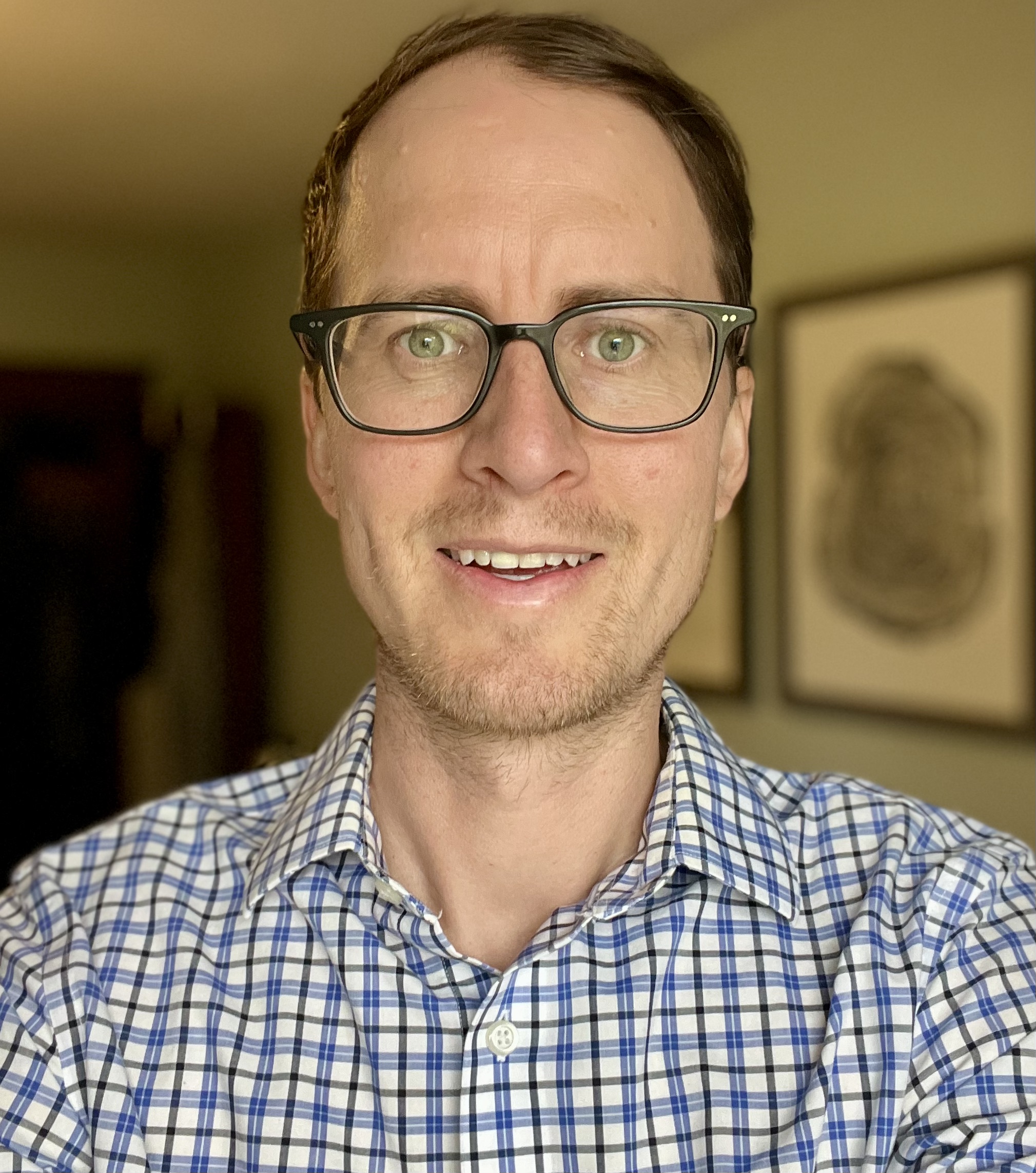}}]{Stephen L. Smith}
(Senior Member, IEEE) received the B.Sc. degree in engineering physics from Queen’s University, Canada, in 2003, the M.A.Sc. degree in electrical and computer engineering from the University of Toronto, Canada, in 2005, and the Ph.D. degree in mechanical engineering from the University of California, Santa Barbara, USA, in 2009. 

He is currently a Professor with the Department of Electrical and Computer Engineering, University of Waterloo, Waterloo, ON, Canada, where he holds a Canada Research Chair in Autonomous Systems and is the Co-Director of the Waterloo Artificial Intelligence Institute. His research interests include control and optimization for autonomous systems, with an emphasis on robotic motion planning and coordination. 

Dr. Smith is an Associate Editor for IEEE TRANSACTIONS ON ROBOTICS and the IEEE OPEN JOURNAL OF CONTROL SYSTEMS. He was the Co-General Chair for the 2021 IEEE International Conference on Robot and Human Interactive Communication
\end{IEEEbiography}
\begin{IEEEbiography}
[{\includegraphics[width=1in,height=1.25in,clip,keepaspectratio]{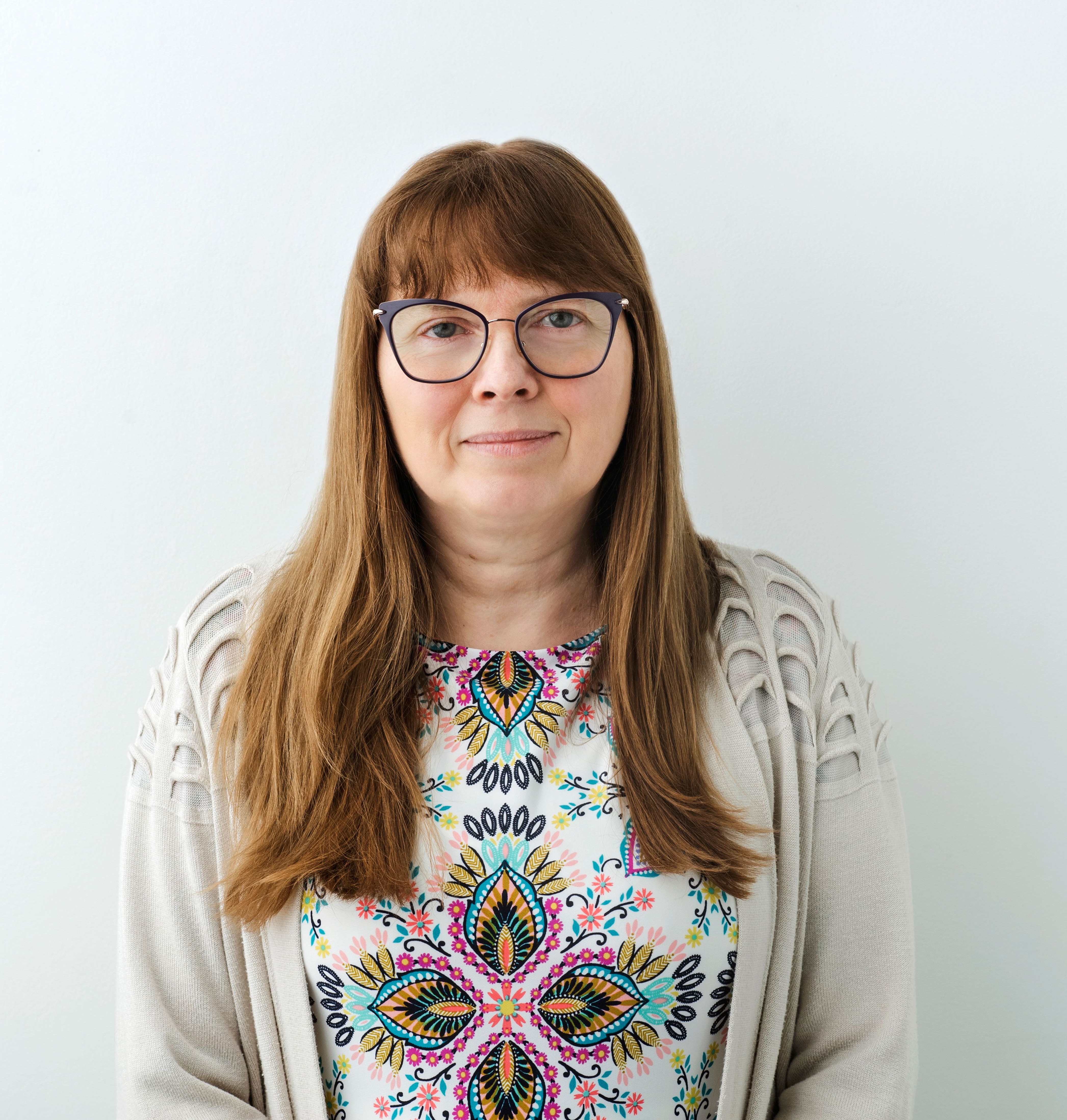}}]{Kerstin Dautenhahn}(Fellow, IEEE) received a Dr. rer. nat in 1993 from University of Bielefeld in Germany, where she previously was awarded a Diploma (equivalent to M.Sc.) in Biology, specialisation in Biological Cybernetic. 

She is currently a Full Professor and Canada 150 Research Chair in Intelligent Robotics with the University of Waterloo, Waterloo, ON, Canada, where she is based in the Department of Electrical and Computer Engineering, and also directs the Social and Intelligent Robotics Research Laboratory. Her research interests include social, intelligent, and cognitive robotics; human–robot interaction; and assistive technology. 

Dr. Dautenhahn is a Fellow of the Royal Society of Canada.
\end{IEEEbiography}

\end{document}